\documentclass[10pt,twocolumn,letterpaper]{article}

\usepackage[pagenumbers]{wacv} 

\usepackage{graphicx}
\usepackage{booktabs}
\usepackage{tikz}
\usepackage{comment}
\usepackage{siunitx}
\usepackage{amsmath,amssymb} 
\makeatletter
\DeclareRobustCommand{\pdot}{\mathbin{\mathpalette\pdot@\relax}}
\newcommand{\pdot@}[2]{%
  \ooalign{%
    $\m@th#1\circ$\cr
    \hidewidth$\m@th#1\cdot$\hidewidth\cr
  }%
}
\makeatother
\usepackage{bbm}
\usepackage{mathtools}
\usepackage{xspace}
\def\bv{\mathbf{v}}
\def\bx{\mathbf{x}}
\def\by{\hat{\mathbf{y}}}
\def\fhat{\hat{f}}
\def\be{\mathbf{e}}

\def\bX{\mathbf{X}}

\def\bR{\mathbb{R}}
\def\mR{\mathcal{R}}
\def\mD{\mathcal{D}}
\def\mN{\mathcal{N}}
\newcommand{\norm}[1]{\left\lVert#1\right\rVert}

\newcommand{\aggexp}{\textsc{Agg$^2$Exp}\xspace}
\usepackage{bbold}
\DeclareMathOperator*{\argmax}{argmax}

\definecolor{myred}{rgb}{0.8554,0.2656,0.2148}
\definecolor{myblue}{rgb}{0.2578,0.5195,0.9531}
\definecolor{myyellow}{rgb}{0.9531,0.7031,0}
\usepackage{booktabs} 
\usepackage{multirow} 
\usepackage[numbers,square,sort&compress]{natbib}

\definecolor{cvprblue}{rgb}{0.21,0.49,0.74}
\usepackage[pagebackref,breaklinks,colorlinks,citecolor=cvprblue]{hyperref}

\usepackage[capitalize]{cleveref}
\crefname{section}{Sec.}{Secs.}
\Crefname{section}{Section}{Sections}
\Crefname{table}{Table}{Tables}
\crefname{table}{Tab.}{Tabs.}

\def\wacvPaperID{850} 
\def\confName{WACV}
\def\confYear{2025}

\begin{document}

\title{Aggregated Attributions for Explanatory Analysis of 3D Segmentation Models}

\author{
Maciej Chrabaszcz\textsuperscript{1,2,}\thanks{Equal contribution. \texttt{maciej.chrabaszcz.dokt@pw.edu.pl} \texttt{h.baniecki@uw.edu.pl}} \quad
Hubert Baniecki\textsuperscript{3,}\footnotemark[1] \quad
Piotr Komorowski\textsuperscript{3} \quad \\
Szymon Płotka\textsuperscript{3,4} \quad
Przemyslaw Biecek\textsuperscript{1,3} \\[.5em]
\textsuperscript{1}Warsaw University of Technology, Poland \quad
\textsuperscript{2}NASK - National Research Institute, Poland \quad \\
\textsuperscript{3}University of Warsaw, Poland \quad
\textsuperscript{4}University of Amsterdam, the Netherlands 
}

\maketitle

\begin{abstract}
Analysis of 3D segmentation models, especially in the context of medical imaging, is often limited to segmentation performance metrics that overlook the crucial aspect of explainability and bias. Currently, effectively explaining these models with saliency maps is challenging due to the high dimensions of input images multiplied by the ever-growing number of segmented class labels. To this end, we introduce \aggexp, a methodology for aggregating fine-grained voxel attributions of the segmentation model's predictions. Unlike classical explanation methods that primarily focus on the local feature attribution, \aggexp enables a more comprehensive global view on the importance of predicted segments in 3D images. Our benchmarking experiments show that gradient-based voxel attributions are more faithful to the model's predictions than perturbation-based explanations. As a concrete use-case, we apply \aggexp to discover knowledge acquired by the Swin UNEt TRansformer model trained on the TotalSegmentator v2 dataset for segmenting anatomical structures in computed tomography medical images. \aggexp facilitates the explanatory analysis of large segmentation models beyond their predictive performance. The source code is publicly available at \url{https://github.com/mi2datalab/agg2exp}.
\end{abstract}


\begin{figure*}
    \centering
    \includegraphics[width=0.9\textwidth]{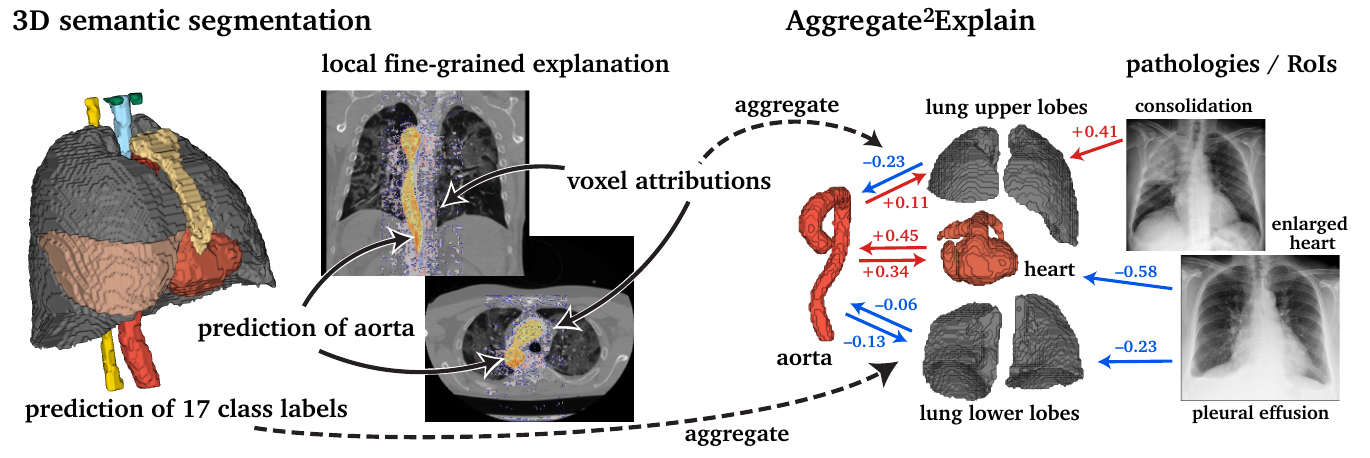}
    \caption{Local fine-grained explanations of 3D semantic segmentation are inherently high-dimensional and incomprehensible to humans. We propose to aggregate voxel attributions over the segmented class labels and other regions of interest (RoIs) to provide a more comprehensive global view on the importance of predicted segments in 3D images. {\color{myred} \textbf{Positive}} and {\color{myblue} \textbf{negative}} attributions between the semantic regions construct an explanatory knowledge graph induced by the black-box segmentation model. \aggexp is especially useful in applied sciences like medical imaging, where explainability enables data bias correction and model validation beyond its predictive performance.}
    \label{fig:abstract}
\end{figure*}

\section{Introduction}

Semantic segmentation of 3D images is a crucial task in computer vision, with useful applications in domains such as robotics~\citep{cheng2021s3net}, autonomous vehicles~\citep{siam2018comparative}, augmented reality~\citep{han2020live}, and healthcare~\citep{li2020shape,he2021dints,hatamizadeh2022unetr,plotka2024swinsmt,farag2024efficientnetsam,szczepanski2024let}.
Notably, automated and accurate segmentation of 3D medical images like computed tomography (CT) scans has the potential to significantly improve the experiences of both clinicians and patients~\citep{rajpurkar2022ai,van2017computational,moor2023foundation,farag2024efficientnetsam}. 
The motivation to aid humans in overcoming diseases~\citep{laslett2012worldwide,ferkol2014global} has lead to continuous improvements of 3D segmentation models~\citep{isensee2020nnunet,li2020shape,he2021dints,jang2021fully,hatamizadeh2022unetr,hatamizadeh2022swinunetr,wasserthal2023totalsegmentator,perera2024segformer3d,jiang2024zept,zhang2024mapseg,chen2024versatile}, which are trained on an ever-increasing amount of medical data~\citep{ma2021abdomenct,luo2022word,ji2022amos,antonelli2022medical,wasserthal2023totalsegmentator,al2023usability}.

Large and complex 3D segmentation models become \emph{black-box} predictive systems, which were initially developed to aid human decision-making.
Unavoidably, the ability to provide understandable, or \emph{explainable}, model outputs is limited and often leads to trust issues~\citep{combi2022manifesto,jin2023guidelines,vandervelden2022explainable}. 
In extreme cases, the lack of understandability conceals biases in data, which negatively influence the model's generalization performance~\citep{hryniewska2021checklist,kim2024discovering}, or even make it discriminate certain groups of patients~\citep{gichoya2022ai}.

To this end, various post-hoc explanation methods have been proposed that aim to approximate the machine learning model's reasoning~\citep{vandervelden2022explainable,zemni2023octet,augustin2024digin,kim2024discovering,sobieski2024global}; the most popular class of methods being \emph{feature attributions}~\citep{simonyan2013deep,ribeiro2016why,lundberg2017unified,sundararajan2017axiomatic,selvaraju2020gradcam,wang2020score,yang2020learning,bora2024slice,jiang2024comparing,achtibatattnlrp}.
The quality of feature attribution methods can be benchmarked with evaluation metrics~\citep{arras2022clevr,hedstrom2023quantus,komorowski2023towards}, including faithfulness to the model~\citep{wu2024faithfulness,wu2024token} and sensitivity to data perturbations~\citep{yeh2019fidelity,bhatt2020evaluating}, which facilitates the work on their improvement.
Recently, researchers proposed pixel attribution methods specific to 2D segmentation models including gradient-based attributions~\citep{hoyer2019grid,chu2021learning}, model-agnostic perturbation-based attributions~\citep{hoyer2019grid,dardouillet2022explainability}, as well as class activation mapping~\citep{vinogradova2020towards,asany2023segxrescam} and layer-wise relevance propagation~\citep{dreyer2023revealing} that are specific to convolutional neural networks.

\emph{What can we infer from these attribution explanations?}
And which of these attribution methods performs best for the semantic segmentation of CT scans? 
It turns out that a manual analysis of 3-dimensional explanations for numerous segmented class labels becomes increasingly challenging. Moreover, no attribution method was proposed or evaluated in the context of explaining 3D segmentation models. To this end, we propose to aggregate voxel attributions that enables explaining the importance of predicted segments and other regions of interest in 3D images (cf.~\cref{fig:abstract}). In summary, our contributions in this paper are as follows:
\begin{enumerate} 
    \item[\textbf{(1)}] We introduce the Aggregate$^2$Explain (\aggexp) methodology for aggregating attributions into segment importance explanations. \aggexp overcomes the inherited limitations of local feature attributions, like high data dimensionality and model complexity, which are specific to the task of 3D image segmentation.
    \item[\textbf{(2)}] We benchmark gradient-based and perturbation-based voxel attributions in explaining 3D segmentation models, which was not done before. A quantitative evaluation based on four criteria: faithfulness, sensitivity, complexity, and efficiency, leads to selecting the best voxel attribution method for our application. 
    \item[\textbf{(3)}] As a concrete use-case, we apply \aggexp to analyze the state-of-the-art model for segmenting anatomical structures in 3D CT scans. \aggexp allows discovering model biases and explaining their abnormal behavior.
\end{enumerate}

\section{Related work}\label{sec:relatedwork}

\subsection{Segmentation of 3D medical images}

Effective 3D segmentation faces computational challenges, data acquisition requirements, and the need for complex algorithms capable of capturing additional spatial dimensions in the 3D space~\citep{hu2021towards}. 
Especially in medical imaging, accurate and explainable 3D segmentation facilitates diagnostics and treatment planning~\citep{jang2021fully,al2023usability,chen2024versatile,jiang2024zept,zhang2024mapseg}. 
A few medical datasets were provided to foster research in this domain, including WORD~\citep{luo2022word}, AMOS~\citep{ji2022amos}, AbdomenCT-1K~\citep{ma2021abdomenct} and Medical Decathlon~\citep{antonelli2022medical} for the segmentation of vital organs inside the abdominal and thoracic body parts. 

More recently, the imperative to construct a foundation model~\citep{moor2023foundation} for a generalizable segmentation of 3D medical images has led to the development of a comprehensive dataset -- TotalSegmentator v2~\citep{wasserthal2023totalsegmentator} -- encompassing the segmented entirety of the human body. 
Thus, we evaluate our explanation methodology on this emerging real-world application of segmenting the anatomical structures within the thorax, which are integral to two of the most common disease-causing human systems, i.e. the cardiovascular and respiratory systems~\citep{laslett2012worldwide,ferkol2014global}. 
We foresee \aggexp to be a general approach applicable to other 3D medical segmentation settings~\citep{isensee2020nnunet,li2020shape,he2021dints,jang2021fully,hatamizadeh2022unetr,hatamizadeh2022swinunetr,wasserthal2023totalsegmentator,perera2024segformer3d,jiang2024zept,zhang2024mapseg,chen2024versatile}.

\subsection{Explanation of (2D) segmentation models}

Unlike for image classification~\citep{combi2022manifesto,vandervelden2022explainable,jin2023guidelines}, only a few works consider explaining 2D segmentation models~\citep{vandervelden2022explainable,zemni2023octet,augustin2024digin,kim2024discovering}. 
The most popular approach we focus on is to measure the feature (pixel) importance and visualize it as an attribution (saliency) map~\citep{ribeiro2016why,sundararajan2017axiomatic,selvaraju2020gradcam,wang2020score}. 
In~\citep{hoyer2019grid}, the first-pixel attribution methods for semantic segmentation are introduced in the form of a saliency grid approximated with gradient-based or perturbation-based algorithms. Follow-up work considers extending gradient-weighted class activation mapping to the segmentation task~\citep{vinogradova2020towards,asany2023segxrescam}.
In~\citep{dardouillet2022explainability}, the authors consider a Shapley-based method that relies on developing an interpretable image representation, e.g. with super-pixels (similar to LIME~\citep{ribeiro2016why}). 
It is noteworthy that pixel-wise attributions can be used to improve segmentation models~\citep{chu2021learning,dreyer2023revealing}. 
More broadly related works consider counterfactual explanations to analyze prediction errors~\citep{zemni2023octet}, concept-based explanations to discover bias~\citep{dreyer2023revealing}, and red-teaming analysis~\citep{jankowski2024redteaming}. Another approach is training inherently interpretable neural networks to overcome the issue of explaining models post-hoc~\citep{sacha2023protoseg,he2023segmentation}. 

None of the above-mentioned works has considered explaining 3D segmentation models, which pose significant algorithmic, computational, and perceptive challenges as the data dimension increases exponentially both in the input and output space. 
We propose an aggregation framework to fill this research gap where explanations of 3D segmentation are visually more comprehensible and thus provide added value in the analysis of medical imaging models specific to 3D segmentation.

\section{\aggexp methodology}\label{sec:methodology}

We consider a 3D semantic segmentation task where voxel attribution scores give an explanation of the segmentation model's prediction. 
Let $f \colon \mathbb{R}^{p} \rightarrow \mathbb{R}^{p \times l}$ be the model function that we wish to explain, where $p \coloneqq width \times height \times depth$ is the flattened 3D spatial dimension and $l$ is the number of segmented class labels.
Let $\bx \in \bR^p$ be an input 3D image and $\by \coloneq f(\bx) \in \bR^{p \times l}$ be the 4D model's output with logit scores for $\bx$.
We denote by $\by_c \in [0, 1]^{p}$ a binary mask that indicates the voxels segmented as a class label $c \in [l] \equiv \{1,\ldots,l\} $, i.e. we have
\begin{equation}\label{eq:prediction_mask}
    \by_c^{(i)} \coloneq
    \begin{cases}
        1, & \text{if} \;\argmax_j \by^{(i,j)} = c, \\
        0, & \text{otherwise}.
    \end{cases}
\end{equation}
We further denote by $f_c(\bx) \colon \mathbb{R}^{p} \rightarrow \mathbb{R}^{p}$ a model function whose output is restricted to logits for class~$c$.

\subsection{Voxel attribution methods for 3D segmentation}\label{sec:methodology_voxel_attribution}

Let $g(\bx; f_c, \cdot)\colon \mathbb{R}^{p} \rightarrow \mathbb{R}^{p}$ denote a general voxel attribution function for input $\bx$, where we emphasize in notation that it explains the model~$f$ predicting class label~$c$, and other method-specific hyperparameters can follow.
Let us observe that the conventional attribution methods are designed for single-dimensional outputs, e.g. a class label, and so they cannot be directly applied to explain $f_c$ that has a $p$-dimensional output for a single class label. 
Therefore, we aggregate $f_c$ to output a single value with an aggregation function $\mathcal{A} \colon \mathbb{R}^{p} \rightarrow \mathbb{R}$, e.g. mean or median.
We denote by $\fhat_c$ an aggregated model function that is explained as a proxy, i.e. we have
\begin{equation}\label{eq:gradient_proxy}
    \fhat_c(\bx) \coloneq \mathcal{A}(f_c(\bx)) = f_c(\bx) \pdot \by_c,
\end{equation} 
where $\pdot$ denotes a dot product in case when $\mathcal{A}$ is a sum aggregation.

\paragraph{Gradient-based attributions.} 
We adapt gradient-based attribution methods~\citep{sundararajan2017axiomatic,hoyer2019grid} to explain differentiable 3D segmentation models using $\fhat_c$ from \cref{eq:gradient_proxy}. 
The vanilla gradient~(VG) attribution is computed by taking a partial derivative of~$\fhat_c$:
\begin{equation}
    g_{\mathrm{VG}}(\bx; \fhat_c) \coloneq \frac{\partial \fhat_c(\bx)}{\partial \bx}.
\end{equation}
SmoothGrad (SG) aggregates VG over a neighborhood around $\bx$ to improve its robustness:
\begin{equation}
    g_{\mathrm{SG}}(\bx; \fhat_c, n, \sigma^2)  \coloneq \frac{1}{n}\sum_{i=1}^{n} g_{\mathrm{VG}}(\bx+ \mathcal{N}(0, \sigma^2); \fhat_c).
\end{equation}
Integrated gradients (IG) aggregate VG on a path from $\bx$ to the selected baseline input $\bx'$ (usually $\bx' = \mathbf{0}$):
\begin{equation}
    g_{\mathrm{IG}}(\bx; \fhat_c, n, \bx')  \coloneq \frac{\bx - \bx'}{n}\sum_{i=1}^{n} g_{\mathrm{VG}}(\bx'+\frac{i}{n}(\bx-\bx'); \fhat_c).
\end{equation}

\paragraph{Perturbation-based attributions.}
Perturbation-based attribution methods are \emph{model-agnostic}~\citep{ribeiro2016why}, i.e. they treat a model as a black-box without assuming any knowledge about its structure, unlike gradient-based methods. 
We adapt KernelSHAP~\citep{lundberg2017unified,dardouillet2022explainability} to explain $\fhat_c$, which requires defining an \emph{interpretable} feature representation.
In case of 3D images, we propose the following two \emph{supervoxel} representations:
\emph{(1)~Semantic:} Given a prediction $\by$ for an input $\bx$, its semantic supervoxel partitioning is defined as $\bv^{(i)} \coloneq \argmax_j \by^{(i,j)}$, where $\bv \in [1, \ldots, l]^{p}$.
\emph{(2)~Cubes:} Given a predefined number of regions $r$, an input $\bx$ is partitioned into supervoxels $\bv \in [1, \ldots, r]^{p}$ defined by even cubes, e.g. an image of size $512^3$ can be partitioned into $64$ regions, each of size $128^3$. 

\begin{figure*}[!t]
    \centering
    \includegraphics[width=0.9\textwidth]{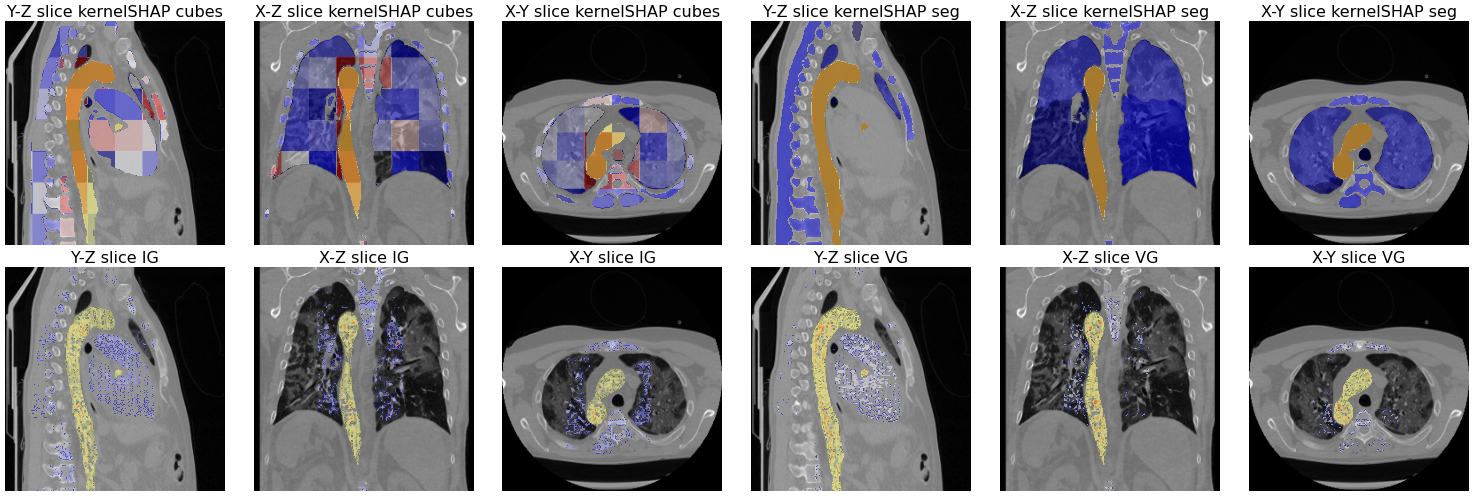}
    \caption{Visualization of voxel attributions explaining the prediction of \emph{aorta} computed with different methods: KernelSHAP (cubes and semantic), IG, and VG. The model's prediction of the \emph{aorta} is highlighted with {\color{myyellow} \textbf{yellow}}. The color mapping of attribution values transitions from positive ({\color{myred}\textbf{red}}) to negative ({\color{myblue}\textbf{blue}}). Particular slices are chosen based on the highest area of the segmented class within each dimension~(X, Y, Z). To improve readability, we display only the top 95\% values for gradient-based methods and remove attribution values for the background segment in KernelSHAP. We show analogous explanations for other class labels and attribution methods in \cref{app:visualization}.}
    \label{fig:example_explanations}
\end{figure*}

Given supervoxel partitioning $\bv$, KernelSHAP function $g_{\mathrm{SHAP}}$ computes the attributions by solving an optimization problem with penalized ridge regression so that
\begin{equation}
    \fhat_c(\bx) \coloneq g_{\mathrm{SHAP}}(\bx; \fhat_c, \bv) \pdot \bx.
\end{equation}
As an illustrative example, we visualize voxel attributions for the 3D segmentation of the class label \emph{aorta} in~\cref{fig:example_explanations}, and provide more such examples in~\cref{app:visualization}.

\subsection{Aggregating attributions for 3D segmentation}\label{sec:methodology_agg2exp}

Surprisingly, our paper is the first to point out that voxel attribution explanations of 3D segmentation have become increasingly complex and impractical (cf.~\cref{fig:example_explanations}). 
For example, given $512^3$ voxels and 17 class labels, one obtains about $10^9$ voxel attributions explaining a single model prediction. 
To overcome this challenge of high complexity stemming from the multiplication of input and output dimensions, we propose to aggregate voxel attributions into simpler explanations that are easier to visualize and analyze. 

Let $\be \in \bR^{p\times l}$ denote an attribution explanation corresponding to a segmentation prediction $f(\bx)$ for an input $\bx$. 
We use $\be_{c}$ to denote attributions for class~$c$, i.e. $\be_c \in \bR^{p}$, and $\be_{c,\mR}$ to denote such an explanation limited to voxels from a region of interest (RoI) encoded with a binary mask $\mR$, i.e. $\be_{c,\mR} = \be_c \pdot \mR$. 

As a natural first choice for RoIs, we consider an aggregating scheme that measures the importance of voxels stemming from a particular class label on other segmented objects, i.e. $\mR_c \coloneq \by_c$. 
To simplify the notation, we often write $\be_{b \rightarrow a}$ instead of $\be_{a,\mathcal{R}_b}$, where $a,b \in [l]$ denote a pair of class labels. 

\paragraph{Local RoI importance.} 
Note that a sparse explanation vector $\be_{b \rightarrow a} \in \bR^{p}$, although it has an interpretation that can be visualized, is still high dimensional. 
In practice, we would like to aggregate $\be_{b \rightarrow a}$ into a single RoI importance value to perform a global model analysis. 
The simplest solution would be to sum or average absolute attribution values $\frac{\norm{\be_{b \rightarrow a}}_1}{p}$, which relates to classic global feature importance~\citep{lundberg2020from}. 
We consider a more reliable approach specific to saliency maps, i.e. the mass accuracy metric~\citep{arras2022clevr}, which measures how much ``mass'' the explanation method gives to voxels within the RoI.
We have
\begin{equation}\label{eq:mass_accuracy}
    e_{b \rightarrow a} \coloneq \frac{\norm{\be_{b \rightarrow a}}_1}{\norm{\be_a}_1},
\end{equation}
where one can also consider aggregating only positive or only negative attributions separately. 
Estimation of \emph{local} RoI importance values for all the pair-wise combinations of class labels constructs a network described by a \emph{local} explanation matrix $\mathcal{E}^{(a,b)}_\bx \coloneqq e_{b \rightarrow a}$, where $\mathcal{E}_\bx \in \bR^{l \times l}$. 

\paragraph{Global RoI importance.}  
Furthermore, we consider analyzing the \emph{global} RoI importance, which is an expected value of local RoI importances aggregated as $\mathcal{E}^{(a,b)}_\bX \coloneq \frac{1}{n} \sum_{\bx \in \bX} \mathcal{E}^{(a,b)}_\bx$, where $\bX$ is a set of $n$ inputs $\bx$. Global $\mathcal{E}_\bX$ and local $\mathcal{E}_\bx$ explanation matrices can be conveniently visualized as a 2D heat map or a directed graph (cf. \cref{fig:abstract}).

To summarize, the \aggexp methodology consists of:
\begin{enumerate}
    \item \emph{Voxel attributions.}
    Compute fine-grained attributions for each voxel per each class label that the model is segmenting using any method described in \cref{sec:methodology_voxel_attribution}.
    \item \emph{Local RoI importance.}
    Aggregate voxel attributions based on a predefined set of RoIs, which include the segmented class labels. This step lowers the complexity of explanations. Note that the set can be extended by other RoIs independent of the model's predictions, e.g. output from another segmentation model or annotated by a human. 
    \item \textit{Global RoI importance.}
    Aggregate local RoI importance with an average over a set of inputs to obtain global importance. This step enables a more comprehensive global view of the 3D segmentation model.
    \item \textit{Visualize and analyze the explanation matrices.}
\end{enumerate}

We acknowledge that using function $\fhat_c(\bx)$ in step 1. may be considered as an initial (third) aggregation step.

\subsection{Eval. metrics for voxel attribution methods}\label{sec:methodology_evaluation_metrics}

Choosing an appropriate voxel attribution method is crucial because it serves as a backbone of \aggexp.
We propose to use the four metrics established for evaluating \emph{feature} attribution methods to select the best \emph{voxel} attribution method for further aggregation. 
In this section, we define the faithfulness, sensitivity, complexity, and efficiency of attribution methods for 3D segmentation, some of which were previously proposed for image classification~\citep{yeh2019fidelity,bhatt2020evaluating,komorowski2023towards,hedstrom2023quantus,wu2024faithfulness,wu2024token} and 2D segmentation~\citep{hoyer2019grid,dreyer2023revealing}.  
An evaluation metric $\mu$ considers an attribution method $g$, an aggregated model function $\fhat_c$ and an input $\bx$, and metric-specific parameters, to yield a scalar value $\mu(g; \fhat_c, \bx, \cdot)$ that quantifies the method's quality. The final value results from an aggregation over a set of class labels and inputs, e.g. a validation set.

\paragraph{Faithfulness to the model.} 
Intuitively, the faithfulness metric measures whether voxels indicated as important by an attribution method are indeed important to the model's prediction. 
The influence is measured by removing a subset of voxels $S \subseteq [p]$ and observing the resulting change in $f_c$.
Let $\bx_{S}$ denote an input $\bx$ with the voxels indexed by $S$ set to the baseline value 0.
We define faithfulness as
\begin{equation}
\begin{split}
    & \mu_{\mathrm{F}}(g; \fhat_c, \bx, n, m) =\\
    & \underset{\substack{S_1,\ldots,S_n \in \binom{[p]}{m}}}{\mathrm{correlation}}\Big(g(\bx; \fhat_c) \pdot \mathbf{1}, \big(\fhat_c(\bx) - \fhat_c(\bx_{S_i})\big) \pdot \mathbf{1} \Big),
\end{split}
\end{equation}
where the correlation is estimated with bootstrapping over $n$ rounds of sampling a subset $S_i$ of size $m$. 
We desire an attribution method $g$ to be highly faithful.

\paragraph{Sensitivity to data perturbation.} 
The sensitivity metric relates to robustness and determines how close attribution explanations are for similar inputs. 
Let $\mD: \mathbb{R}^p \times \mathbb{R}^p \rightarrow \mathbb{R}$ denote a given distance metric between two attribution explanations. 
We define sensitivity as
\begin{equation}
\begin{split}
    & \mu_{\mathrm{S}}(g; \fhat_c, \bx, n, \sigma^2) =\\
    & \frac{1}{n}\sum_{i=1}^{n} \mD\big(g(\bx; \fhat_c), g(\bx + \mathcal{N}(0, \sigma^2); \fhat_c)\big),
\end{split}
\end{equation}
where $\mN(0, \sigma^2)$ denotes Gaussian noise, and its addition to an input~$\bx$ effectively defines a $\sigma$-neighborhood around~$\bx$.
We desire an attribution method $g$ to have low sensitivity.

\paragraph{Complexity (sparsity).} 
In general, a lower complexity of explanations is desirable as it improves their comprehensibility by focusing on fewer, more influential features~\cite{poursabzi2021manipulating}. 
To quantify the complexity as a measure of the explanation's sparsity, we calculate the fraction of voxels whose attribution scores exceed a specified threshold~$\theta$, i.e.
\begin{equation}
    \mu_{\mathrm{C}}(g; \fhat_c, \bx, \theta) = \frac{\big|\{i \colon g(\bx; \fhat_c)^{(i)} > \theta \}\big|}{p}.
\end{equation}

\paragraph{Computational efficiency.} 
Finally, we measure the time in seconds required to compute attributions using method~$g$.

\section{Experimental analysis}\label{sec:experiments}

In experiments, we first benchmark voxel attribution methods to motivate the choice of a backbone for \aggexp~(\cref{sec:experiments_benchmark}), and then describe in detail a use-case of \aggexp in our application~(\cref{sec:experiments_usecase}).

\subsection{Setup}

\paragraph{Datasets.} 
To evaluate \aggexp, we use two datasets: the publicly available TotalSegmentator v2~\citep[TSV2,][]{wasserthal2023totalsegmentator} test set, and our private dataset~(denoted as B50),\footnote{An ethical approval was granted by the Ethics Committee of the \emph{Medical University of Warsaw}.} 
comprising 32 and 127 CT scans, respectively. 
The inclusion of the latter is motivated by our real-world application of \aggexp in practice to analyze the 3D segmentation model that will potentially be deployed in hospital software.
Both datasets consist of 3D CT scans of humans' chests, which include healthy (normal) cases, as well as cases with various thoracic pathologies indicating illness. 
Following~\citep{wasserthal2023totalsegmentator}, we select 16 of the most important anatomical structures of the thorax as the class labels for segmentation, which, including background, results in a set of 17 labels in total. 
Note that the above-mentioned CT scans have all those classes present. 
As a preprocessing step, we standardize CT scans to an isotropic resolution of 1.5 $\times$ 1.5 $\times$ 1.5 $\mathrm{mm}^3$, clip Hounsfield units to the range $[-1024, 1024]$, and then scale them linearly to $[0, 1]$.

\paragraph{Model.} 
We use the well-established Swin UNETR~\citep{hatamizadeh2022swinunetr,tang2022self} architecture as a baseline model for our experiments. 
It is a faster\footnote{Model inference performs about $10\times$ faster due to changes in model size and architecture.} 
and more accessible\footnote{TotalSegmentator is closed behind a complex Python API, which, for example, prevents its accessible differentiation to approximate gradients.} 
version of the original TotalSegmentator model~\citep{wasserthal2023totalsegmentator} that is based on the older nnUNet~\citep{isensee2020nnunet} architecture.
We train the model on the TSV2 train set. We observe performance improvements specifically due to merging multiple ribs and vertebrae into a single class, as well as carefully selecting the 16 anatomical structures. 
Overall, it lowers the computational complexity while allowing us to segment anatomical structures vital for our collaborating radiologists. 

\paragraph{Reproducibility.} Both the private dataset B50 and the network weights of our trained model are available upon a reasonable request. We provide further details on the hyperparameters of the voxel attribution methods and their evaluation metrics in~\cref{app:hyperparameters}.

\subsection{Benchmarking voxel attribution methods}\label{sec:experiments_benchmark}

\begin{table*}[ht]
    \centering
    \caption{Benchmark of 4 voxel attribution methods (including KernelSHAP in the two variants) with 4 evaluation metrics on 2 validation datasets. We report means and standard deviations over a set of inputs and class labels, and underline the best scores in each metric.}
    \label{tab:benchmark}
    {\small
    \begin{tabular}{ll|cccc}
    \toprule
        \textbf{Dataset} & \textbf{Attribution method} $g$ & \textbf{Faithfulness} $\mu_{\mathrm{F}} \uparrow$ & \textbf{Sensitivity} $\mu_{\mathrm{S}} \downarrow$ & \textbf{Complexity} $\mu_{\mathrm{C}} \downarrow$ & \textbf{Efficiency [s]} $\downarrow$ \\
    \midrule
        \multirow{5}{*}{TSV2} & KernelSHAP (cubes) & $0.038_{\pm 0.11}$ & \underline{$0.157_{\pm 0.10}$} & $0.525_{\pm 0.23}$ & $4278_{\pm 3013}$ \\
        & KernelSHAP (semantic) & $0.158_{\pm 0.21}$ & $0.346_{\pm 0.49}$ & $0.924_{\pm 0.05}$ & $857_{\pm 604}$ \\
        & Vanilla Gradient & \underline{$0.383_{\pm 0.14}$} & $1.130_{\pm 0.26}$ & \underline{$0.001_{\pm 0.00}$} & \underline{$12_{\pm 7}$} \\
        & Integrated Gradients & $0.213_{\pm 0.13}$ & $0.848_{\pm 0.23}$ & \underline{$0.001_{\pm 0.00}$} & $210_{\pm 148}$ \\
        & SmoothGrad & $0.331_{\pm 0.14}$ & $0.773_{\pm 0.20}$ & \underline{$0.001_{\pm 0.00}$} & $209_{\pm 148}$ \\
    \midrule
        \multirow{5}{*}{B50} & KernelSHAP (cubes) & $0.009_{\pm 0.11}$ & $0.341_{\pm 0.38}$ & $0.681_{\pm 0.11}$ & $10427_{\pm 1751}$ \\
        & KernelSHAP (semantic) & $0.040_{\pm 0.11}$ & \underline{$0.181_{\pm 0.29}$} & $0.918_{\pm 0.04}$ & $2087_{\pm 349}$ \\
        & Vanilla Gradient & \underline{$0.411_{\pm 0.14}$} & $1.244_{\pm 0.33}$ & \underline{$0.000_{\pm 0.00}$} & \underline{$27_{\pm 5}$}\\
        & Integrated Gradients & $0.248_{\pm 0.12}$ & $0.943_{\pm 0.21}$ & \underline{$0.000_{\pm 0.00}$} & $510_{\pm 86}$ \\
        & SmoothGrad & $0.309_{\pm 0.13}$ & $0.874_{\pm 0.21}$ & \underline{$0.000_{\pm 0.00}$} & $509_{\pm 86}$ \\
    \bottomrule
    \end{tabular}
    }
\end{table*}

We benchmark voxel attribution methods introduced in~\cref{sec:methodology_voxel_attribution} with evaluation metrics introduced in~\cref{sec:methodology_evaluation_metrics} in our setup.
Table~\ref{tab:benchmark} shows results where we report a mean of the metrics' values over all patient cases and seven class labels,\footnote{Class labels: \emph{background, aorta, lung lower lobe left, lung lower lobe right, trachea, heart, ribs}.} which is due to high computational complexity of calculating those metrics, i.e. this benchmark took around two weeks to compute on an 8$\times$A100 GPU node, highlighting the particular complexity of this task. 
KernelSHAP is estimated based on the two introduced supervoxel representations, where \emph{semantic} considers all 17 class labels, and \emph{cubes} is based on about $64$ regions, each of size $128^3$, depending on the size of a scan.
In general, the two KernelSHAP variants are significantly less efficient than gradient-based methods because they need to perform a large number of model inference steps. 
Gradient-based attribution methods have lower complexity than the perturbation-based KernelSHAP because the latter attributes larger parts of the input as important.

To conclude, we choose the SmoothGrad voxel attribution method for \aggexp in our use-case because it performs better in terms of faithfulness and sensitivity than Integrated Gradients, at the same time providing a reasonable faithfulness to sensitivity ratio as compared to the unfaithful KernelSHAP and oversensitive Vanilla Gradient.

\subsection{Use-case: Explanatory analysis of a model for segmenting anatomical structures in CT scans} \label{sec:experiments_usecase}

We aim to analyze the complex model's behavior considering machine learning and domain knowledge. 

\begin{figure}[t]
    \centering
    \includegraphics[width=0.9\columnwidth]{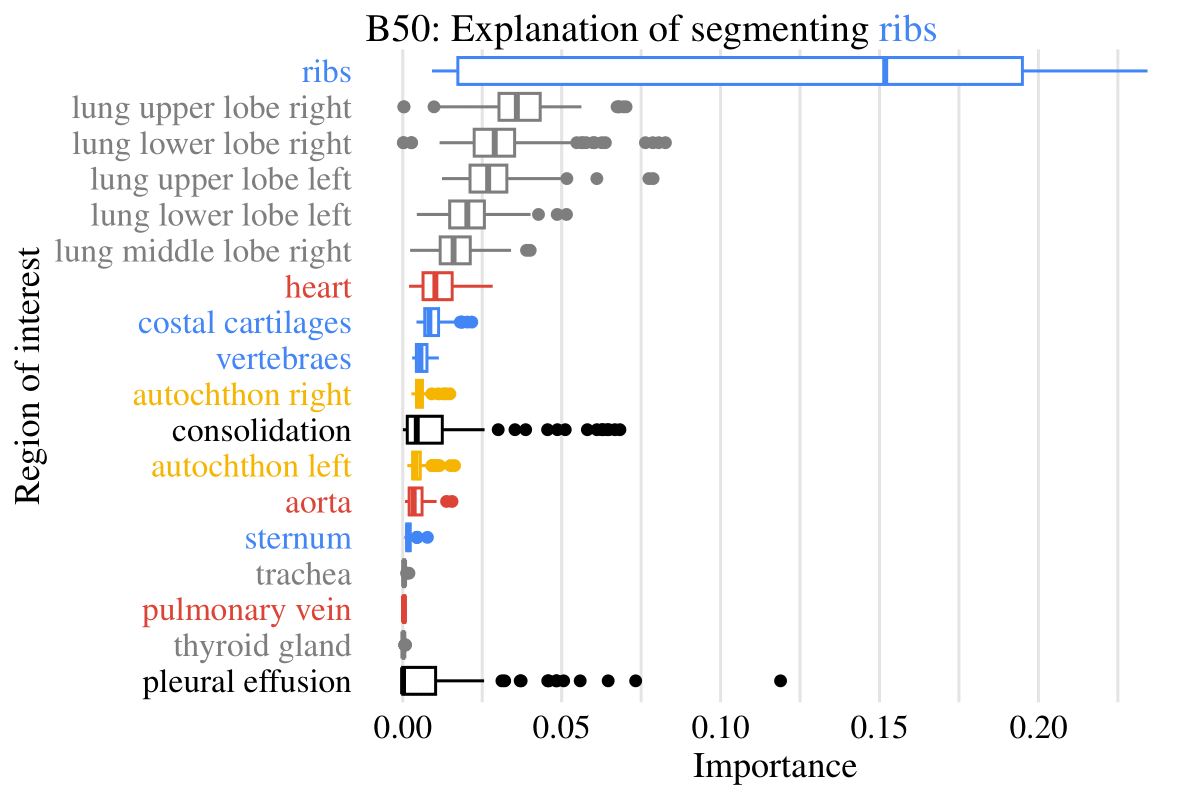}
    \includegraphics[width=0.9\columnwidth]{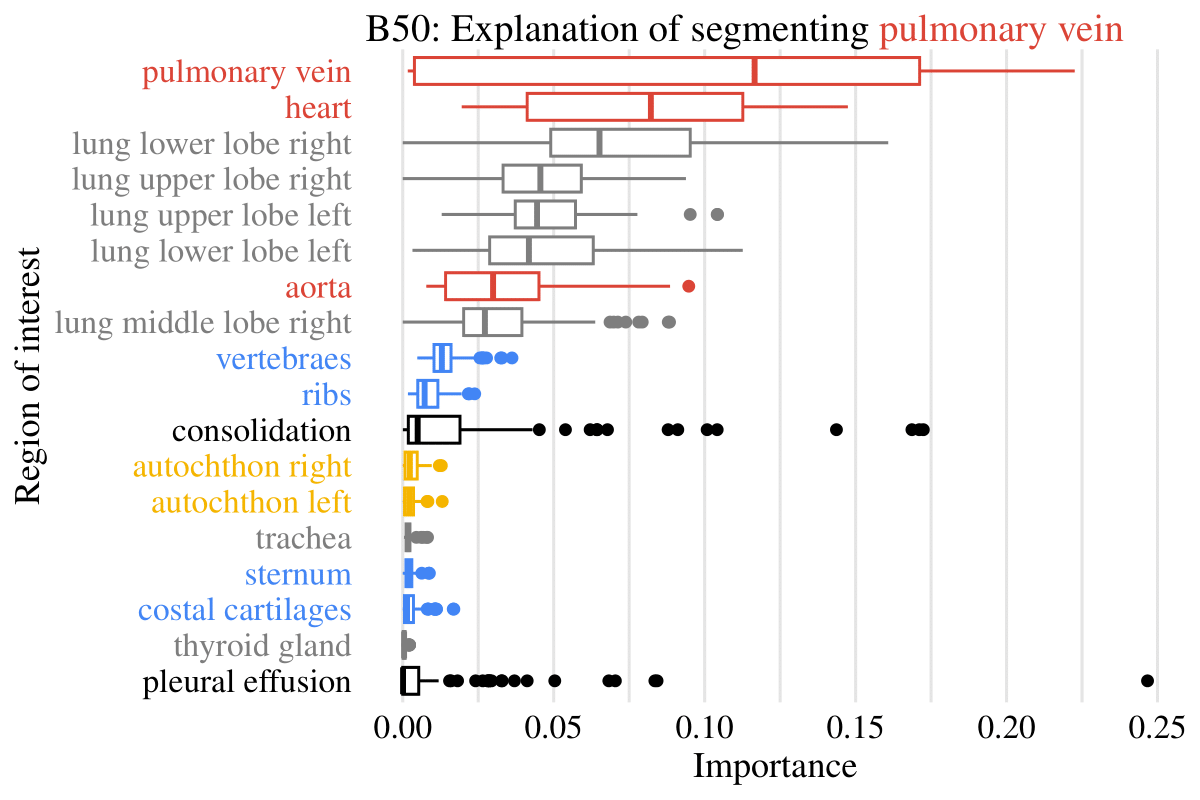}
    \caption{Distribution of local RoI importances for two class labels: \emph{ribs} and \emph{pulmonary vein}. We color objects by their semantic meaning, i.e., cardiovascular system in {\color{myred} \textbf{red}}, muscles in {\color{myyellow} \textbf{yellow}}, bones in {\color{myblue} \textbf{blue}}, other organs in {\color{darkgray} \textbf{grey}}, and lung pathologies in \textbf{black}.}
    \label{fig:glocal_b50}
\end{figure}

\cref{fig:glocal_b50} shows the distribution of local RoI importances for two exemplary class labels: \emph{ribs} and \emph{pulmonary vein}. 
Following~\citep{wasserthal2023totalsegmentator}, we color objects by their semantic meaning, i.e., the cardiovascular system in {\color{myred} \textbf{red}}, muscles in {\color{myyellow} \textbf{yellow}}, bones in {\color{myblue} \textbf{blue}}, other organs in {\color{darkgray} \textbf{grey}}, and lung pathologies in \textbf{black}. 
Masks for the latter are available in our B50 dataset, which allows us to extend the analysis to regions that are of interest for the segmentation of chest CT scans.
The model segments \emph{ribs} by relying mostly on the part of input classified as \emph{ribs} itself. Contrary, segmenting a \emph{pulmonary vein} requires the model to use a broader context, i.e. the regions of \emph{heart} and \emph{lungs} highly contribute to \emph{pulmonary vein} segmentation.
Crucially, for many input cases (outliers denoted with black dots), regions of the \emph{consolidation} and \emph{pleural effusion} pathologies influence the model's predictions.

\begin{figure}[!t]
    \centering
    \includegraphics[width=0.99\columnwidth]{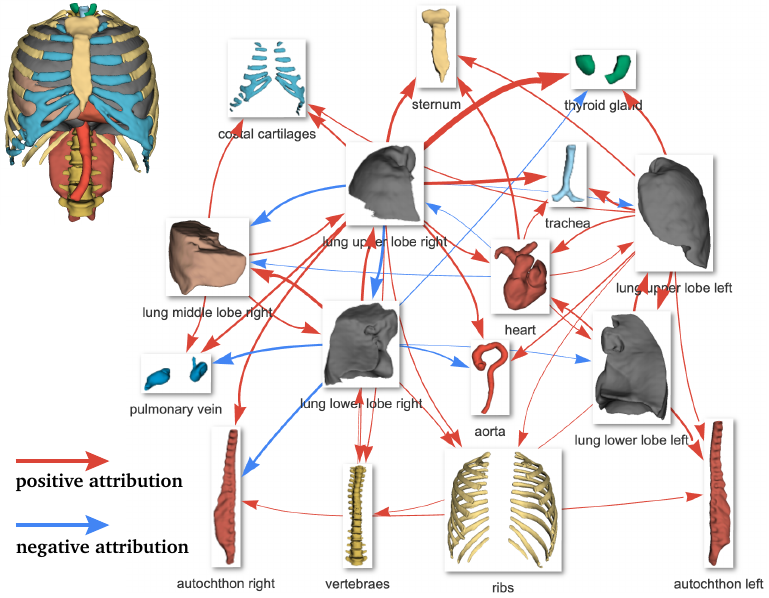}
    \caption{A global RoI importance explanation of the model's 3D segmentation. Aggregated attributions provide a measure of semantic importance between the segmented objects in TSV2. We visualize about one-third of the most important edges for clarity.} 
    \label{fig:graph_tsv2}
\end{figure}

\cref{fig:graph_tsv2} shows a semantic graph of global RoI importances between the segmented objects based on the TSV2 dataset. 
Note that the graph is complete, but we visualize its simplification where Top-3 importances contributing to each node are shown. 
\aggexp shows which anatomical structures contribute to segmenting each class label, and what are the main physical dependencies in the model. 
It is possible to effectively visualize and analyze aggregated explanations, which is not the case for high-dimensional voxel attributions (cf. \cref{fig:example_explanations}).

\paragraph{Analysis of outliers with \aggexp.}

We use \aggexp to perform local outlier analysis, i.e. find patient cases for which the model behaves abnormally. 
We apply the rather straightforward Isolation Forest \citep[IF,][]{Liu2012IsolationBasedAD} algorithm to the matrix consisting of local RoI importances.
We do that for each of the 17 class labels separately, resulting in an IF trained on 17 features from \aggexp per each class label.
We first train IF on the TSV2 training set in order to model the normal behavior and then apply it to validation data (in this case our B50 dataset). 

To answer whether \emph{the model behaves differently for certain groups of patients}, we use their metadata including characteristics like sex and age.
We found that there are no significant differences in the sex and age distributions between inliers and outliers, which suggests that the potentially abnormal behavior of the model is not related to certain subgroups of patients.

\begin{figure*}[t]
    \centering
    \includegraphics[width=0.9\textwidth]{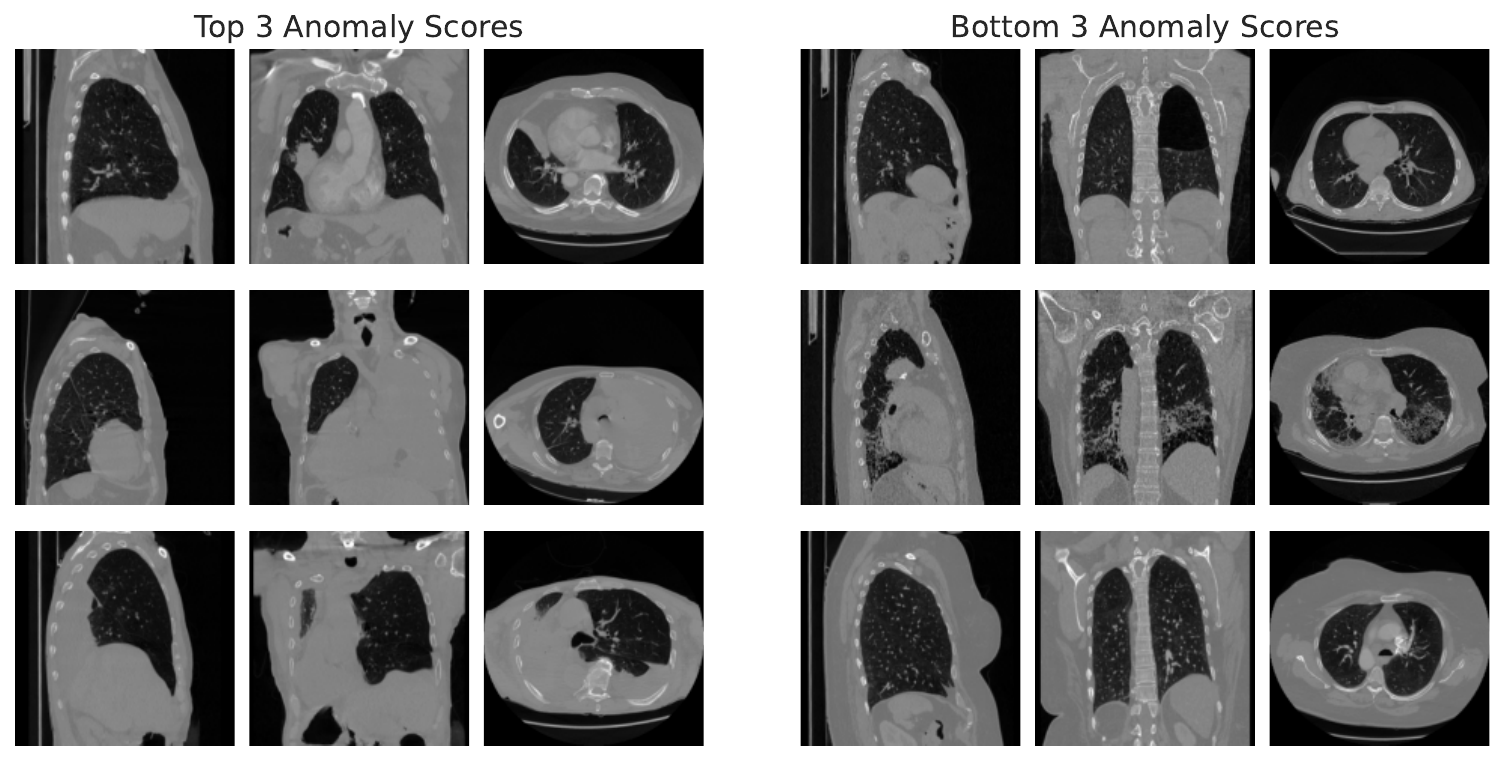}
    \caption{Comparison between the three patient cases with the highest (\textbf{left}) and lowest (\textbf{right}) anomaly scores in our B50 dataset. 
    }
    \label{fig:top3_outliers_inliers}
\end{figure*}

Having access to textual reports of CT scans prepared by radiologists, we analyze examples tagged as outliers by IF trained on \emph{lung lower lobe right} \aggexp features.
\cref{fig:top3_outliers_inliers} shows CT scan examples with the lowest and highest anomaly scores. 
We notice that patients with higher anomaly scores have had profound changes near or inside the right lung. 
This suggests that the model works abnormally when segmenting \emph{lung lower lobe right} for those types of patients, and it could be beneficial to add more such examples to training data for better generalization. 

Additionally, we analyze whether the abnormal model's behavior entails worse predictive performance (cf.\cref{app:outlier}). 
We conduct Spearman's rank test between the anomaly score returned by IFs and the DSC from TSV2 test data for each class label, which shows a statistically significant correlation for some labels (including \emph{lungs}). 
It further suggests that the aggregated attributions can highlight weak model performance for specific labels in inference time, i.e. without the ground truth segmentation masks.

We also conduct a Spearman's rank correlation test to analyze the relationship between anomaly scores averaged over labels and the mean DSC. The test returns a p-value of $\approx0.0024$, which shows a statistically significant relation between the mean anomaly score and the mean DSC.

\paragraph{\aggexp: How crucial is the context?}

\begin{figure}
    \centering
    \includegraphics[width=0.9\columnwidth]{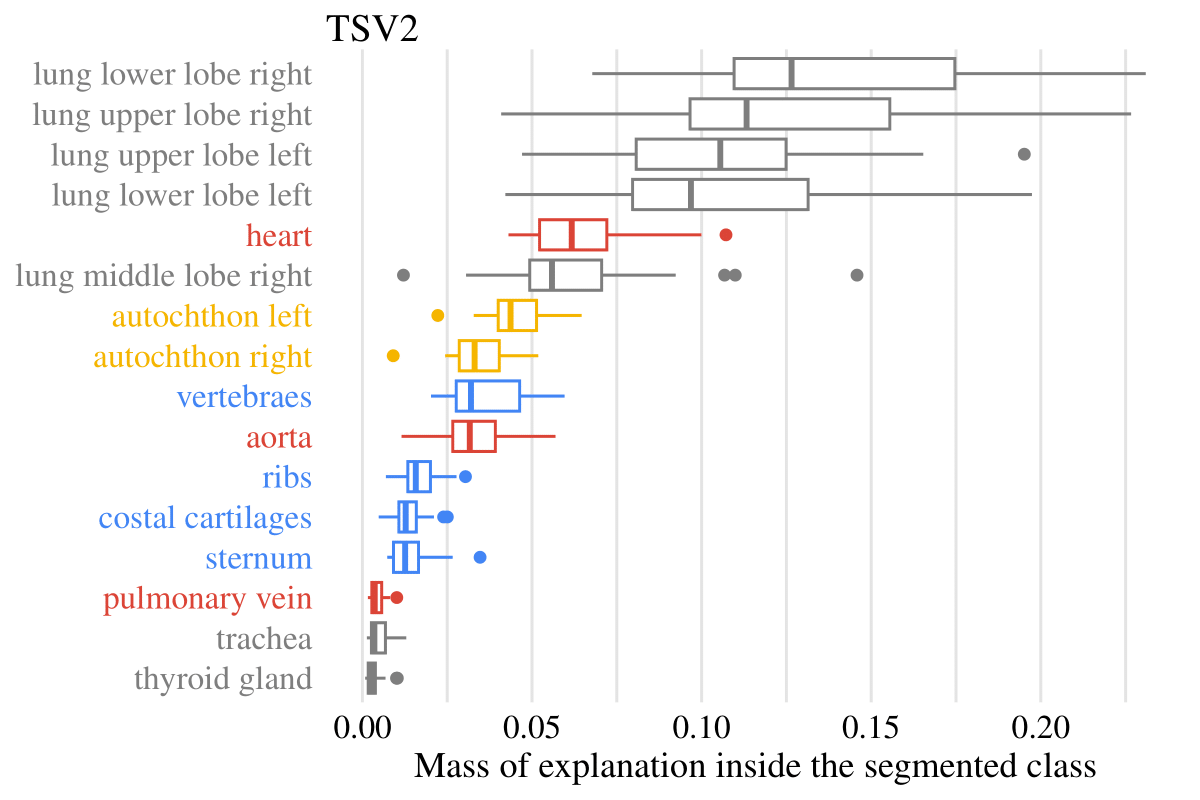}
    \includegraphics[width=0.9\columnwidth]{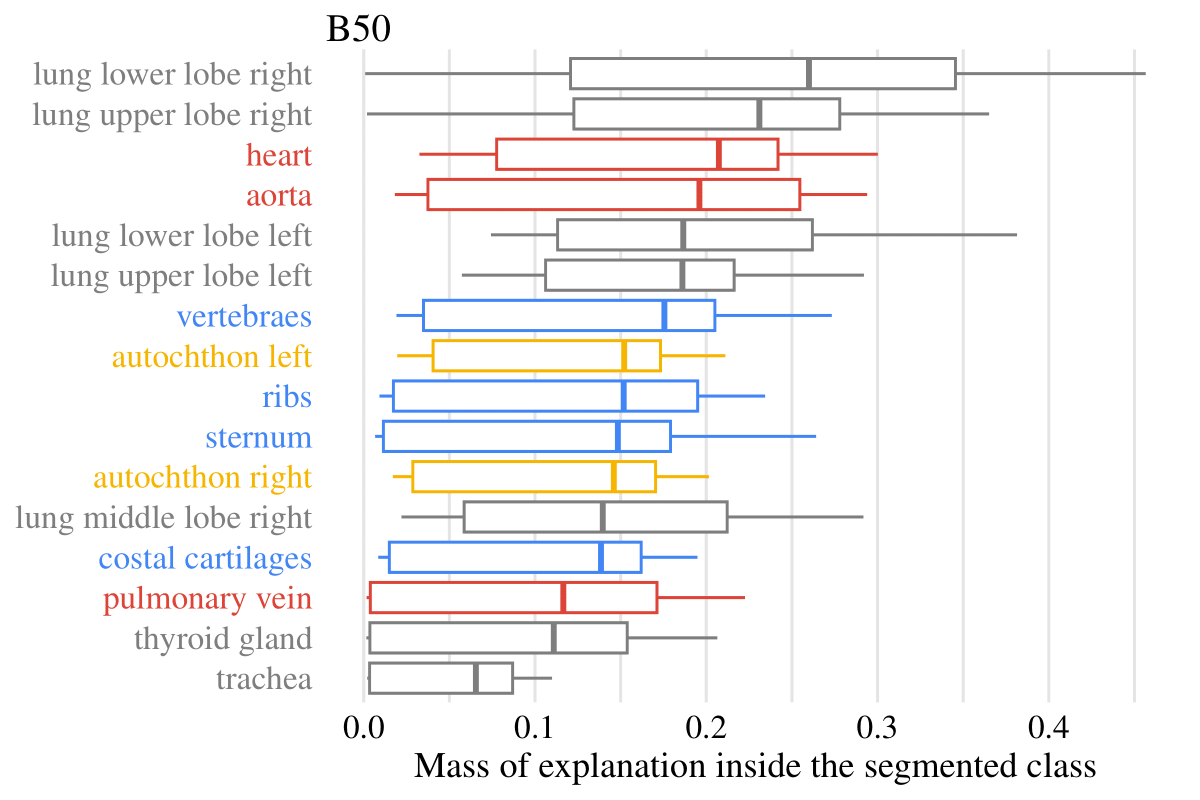}
    \caption{Distribution of the attribution mass lying inside the explained segmented class. Results differ between the two datasets.
    }
    \label{fig:context_used_per_class}
\end{figure}

Finally, we check how much \emph{context} the model uses to segment each class. 
We measure the mass of attributions \cref{eq:mass_accuracy} inside the segmented class to inspect this property, which allows us to understand better how much contextual information the model uses for the segmentation of each class. 
As observed in~\cref{fig:context_used_per_class}, the model uses the least contextual information for the segmentation of \emph{trachea} and \emph{thyroid gland}, whereas it uses nearly only contextual information for the segmentation of \emph{lungs}.

\paragraph{Conclusion.} To summarise, \aggexp provides key insights into the segmentation model: 
\begin{enumerate}
    \item There are patient outliers in the space of explanations.
    \item Segmenting an object relies on the context of physically and semantically nearby objects.
    \item The model behaves differently on our private out-of-distribution dataset, which raises concerns.
\end{enumerate}
We discuss further analyses and provide complementary explanations for the two datasets in~\cref{app:explanations}.

\section{Limitations and future work}

We identify three main limitations of \aggexp that provide natural opportunities for future work. By design, \aggexp inherits the general limitations of feature attribution methods~\citep{rudin2019stop,jin2023guidelines}, which at best approximate an explanation of the model's predictions. As a countermeasure, we proposed to perform a benchmark considering several desiderata to guide a choice of the most suitable voxel attribution method for a given applied use-case. Certainly, improvements in the faithfulness and efficiency of voxel attribution approximation will lead to improvements of \aggexp~\citep{dreyer2023revealing,wu2024faithfulness}. Second, our proposition of using an aggregated model function~(\cref{eq:gradient_proxy}) as a proxy for an explanation could be suboptimal. 
Although it was proven sufficient for our use-case, future work can consider alternative aggregation measures to achieve different goals. Finally, we acknowledge that \aggexp relates to human-computer interaction in that explanations can be misunderstood by users~\citep{baniecki2023grammar} or even manipulate their perception~\citep{poursabzi2021manipulating}.
Following the arguments posited in~\citep{biecek2024position}, we believe \aggexp to answer the needs of machine learning developers and radiologists debugging our segmentation model, leaving an evaluation via user studies as a future research direction~\citep{rong2024towards}.

\section{Conclusion}

As 3D segmentation models are increasingly deployed in practical applications, it becomes pivotal to ensure that they are safe and unbiased.
To this end, we contributed \aggexp, a methodology for explanatory analysis of 3D segmentation models that includes the definition of voxel attributions, evaluation metrics for their approximators, and provides an intuitive scheme for aggregating attributions into meaningful explanations.
We applied \aggexp to analyze a truly black-box model for segmenting anatomical structures in chest CT scans, which allows discovering concepts learned by the model, its reliance on context, and highlights abnormal patient cases. In fact, we foresee that \aggexp could be applied to analyze 3D segmentation models beyond medical imaging.

\section*{Acknowledgments}
This work was financially supported by the Polish National Center for Research and Development grant number INFOSTRATEG-I/0022/2021-00. 

{\small
\bibliographystyle{ieee_fullname}
\bibliography{references}
}

\clearpage
\appendix
\onecolumn

\section*{Supplementary material for the paper:\\``Aggregated Attributions for Explanatory Analysis of 3D Segmentation Models''}

\section{Visualization of exemplary local explanations for different voxel attribution methods} \label{app:visualization}

\begin{figure*}[ht]
    \centering
    \includegraphics[width=0.8\textwidth]{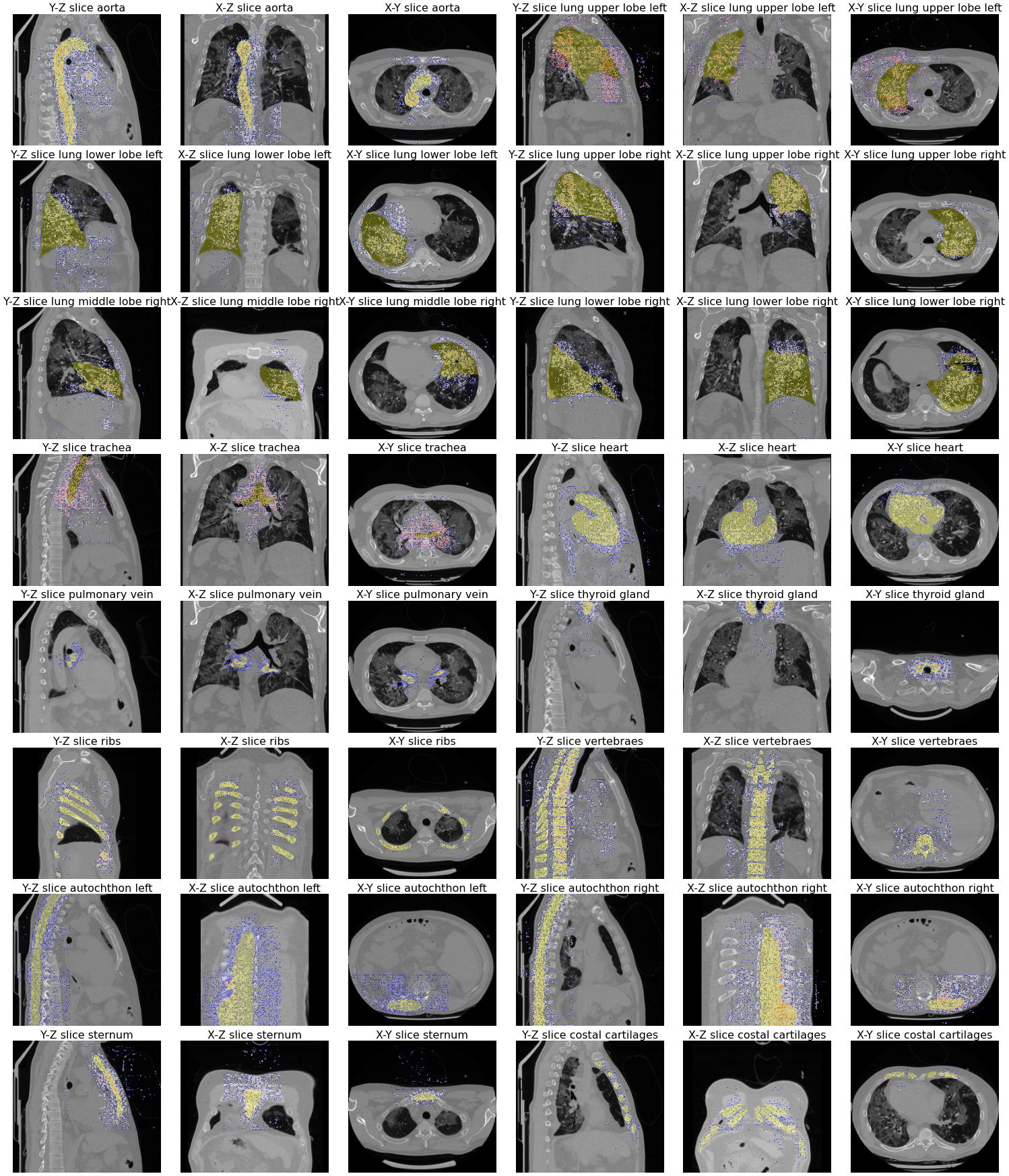}
    \caption{Visualization of VG attribution map across segmentation classes in three dimensions. The color mapping of attributions transitions from blue to red. Slices are chosen based on the highest area of segmented class within each dimension, with the top 95\% values displayed for improved readability. Model predictions for individual organs are highlighted in yellow.}
    \label{fig:example_vg}
\end{figure*}

\begin{figure*}[ht]
    \centering
    \includegraphics[width=0.8\textwidth]{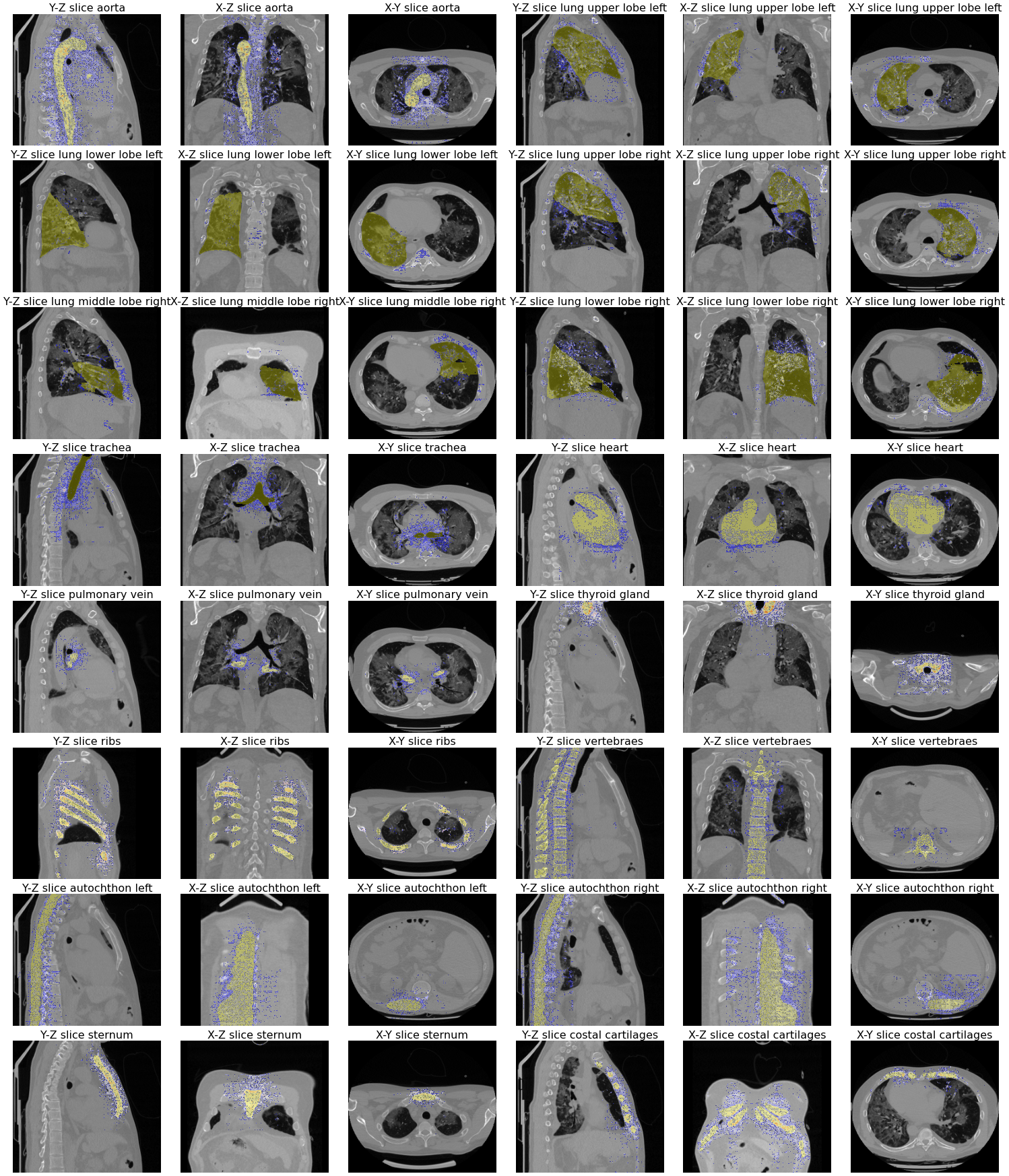}
    \caption{Visualization of IG attribution map across segmentation classes in three dimensions. The color mapping of attributions transitions from blue to red. Slices are chosen based on the highest area of segmented class within each dimension, with the top 95\% values displayed for improved readability. Model predictions for individual organs are highlighted in yellow.}
    \label{fig:example_ig}
\end{figure*}

\begin{figure*}[ht]
    \centering
    \includegraphics[width=0.8\textwidth]{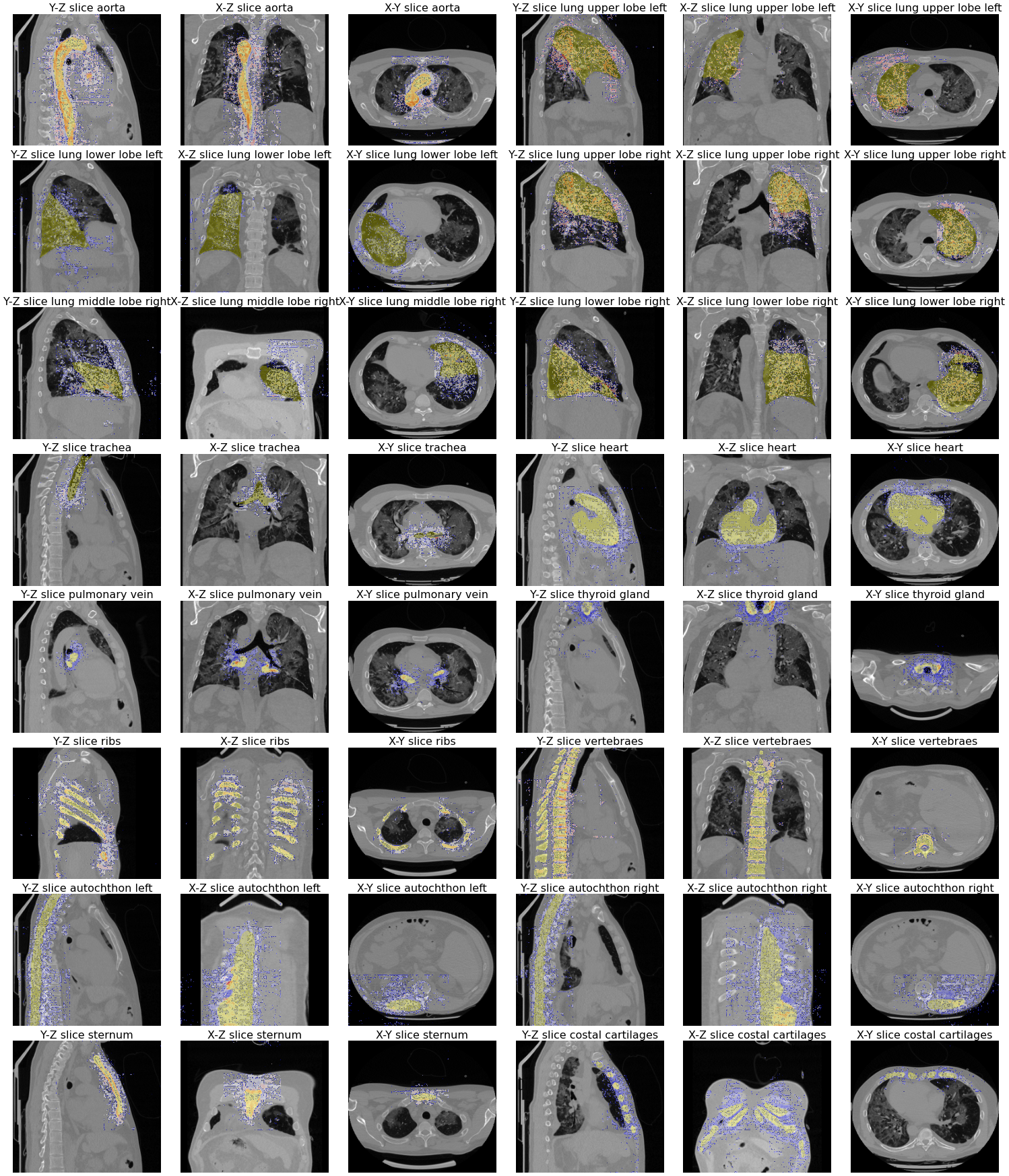}
    \caption{Visualization of SG attribution map across segmentation classes in three dimensions. The color mapping of attributions transitions from blue to red. Slices are chosen based on the highest area of segmented class within each dimension, with the top 95\% values displayed for improved readability. Model predictions for individual organs are highlighted in yellow.}
    \label{fig:example_sg}
\end{figure*}

\begin{figure*}[ht]
    \centering
    \includegraphics[width=0.8\textwidth]{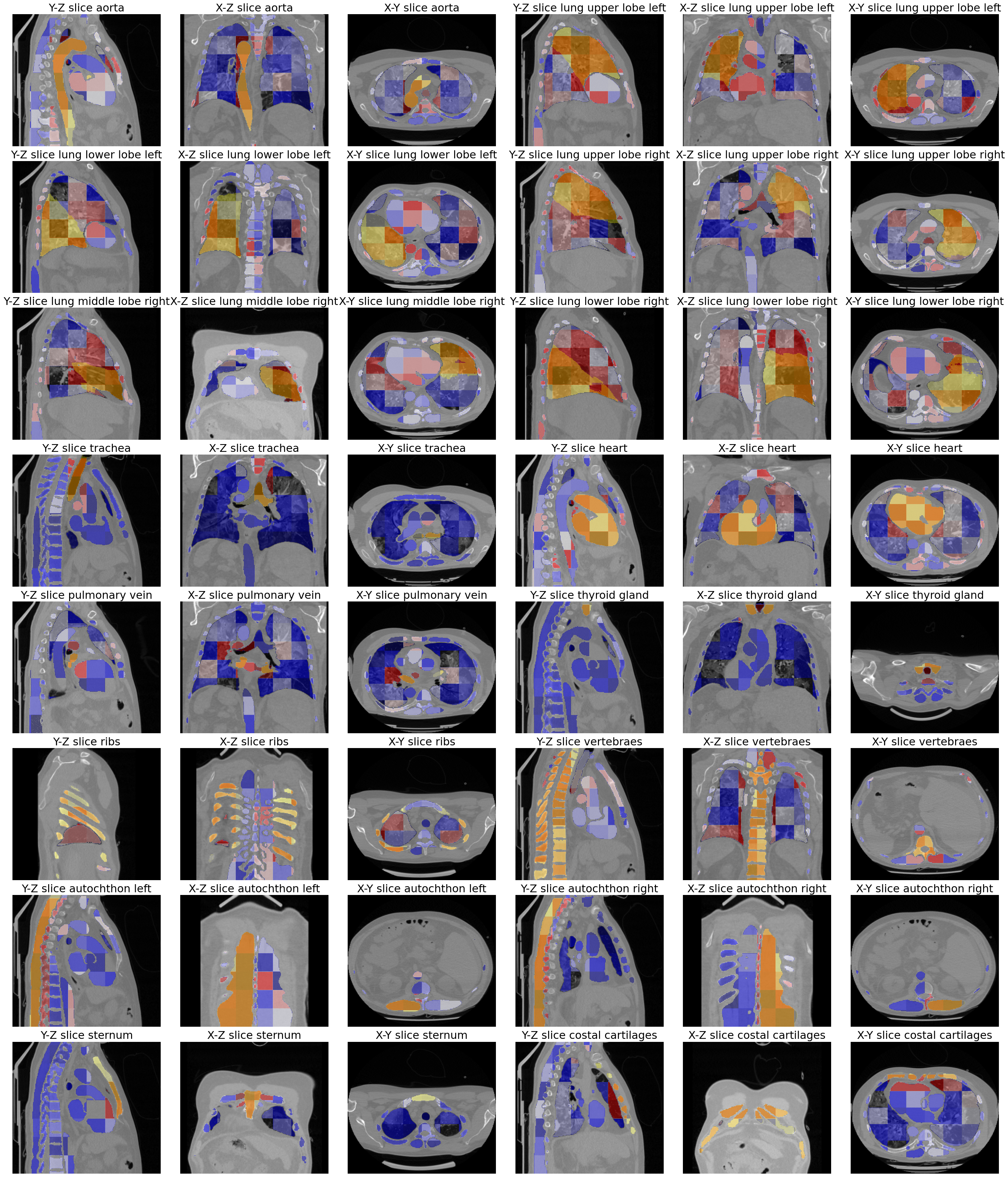}
    \caption{Visualization of KernelSHAP (cubes) attribution map across segmentation classes in three dimensions. The color mapping of attributions transitions from blue to red, attributions are zeroed when they overlap with the background class for better readability. Slices are chosen based on the highest area of segmented class within each dimension. Model predictions for individual organs are highlighted in yellow.}
    \label{fig:example_kernelshap_cubes}
\end{figure*}

\begin{figure*}[ht]
    \centering
    \includegraphics[width=0.8\textwidth]{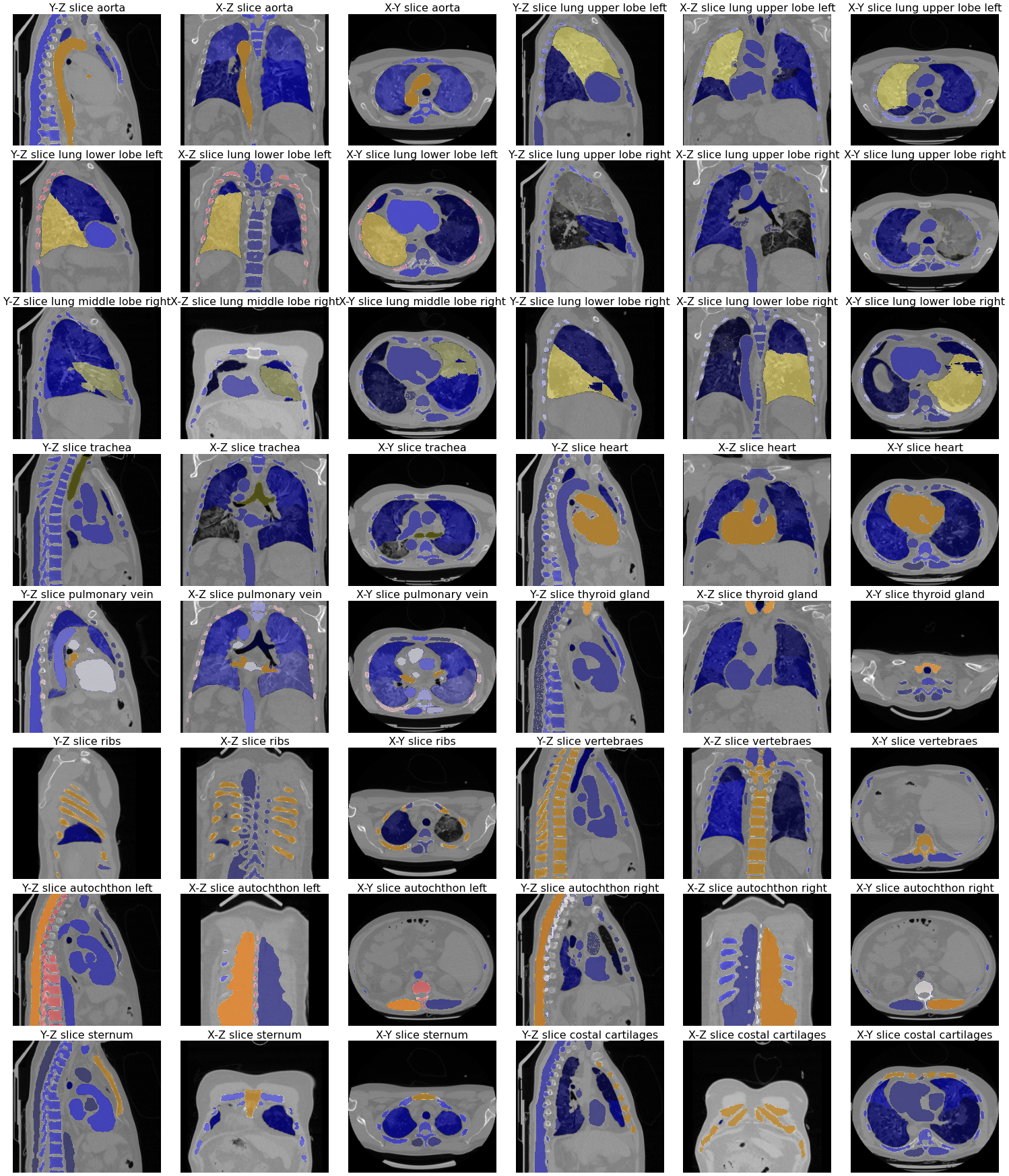}
    \caption{Visualization of KernelSHAP (semantic) attribution map across segmentation classes in three dimensions. The color mapping of attributions transitions from blue to red, attributions are zeroed when they overlap with background class for better readability. Slices are chosen based on the highest area of segmented class within each dimension. Model predictions for individual organs are highlighted in yellow.}
    \label{fig:example_kernelshap_segmentations}
\end{figure*}

See~\cref{fig:example_vg,fig:example_ig,fig:example_sg,fig:example_kernelshap_cubes,fig:example_kernelshap_segmentations}.

\clearpage
\section{Experimental setup: hyperparameters} \label{app:hyperparameters}

\subsection{Attribution methods parameters}
\begin{itemize}
    \item \textbf{Integrated Gradients} We used $x'=0$ as baseline and $n=20$ while calculating IG attribution.
    \item \textbf{SmoothGrad} We used $n=20$ and gaussian noise with $\sigma = 0.1 \cdot (\max(x)-\min(x))$.
    \item \textbf{Kernel SHAP (cubes)} We used 512 components for cubes and 1000 samples for attribution calculation. We replace selected features (cubes) with zeros.
    \item \textbf{Kernel SHAP (semantic)} We used 200 samples and replaced features (segmentation regions) with zeros.
\end{itemize}
\subsection{Attributions Evaluations metrics parameters}
When calculating each metric we normalized attributions into $[0,1]$.
\begin{itemize}
    \item \textbf{Faithfulness} We used subset size $S=224^2$. We used Pearson's Correlation as a similarity function. We replaced selected regions with zeros. We used $n=100$.
    \item \textbf{Sensitivity} For sensitivity, we used $n=3$ due to the computational complexity of Kernel SHAP attributions and a lower bound of $0.1$.
    \item \textbf{Complexity} When calculating Complexity we used $\theta = 0.1$.
\end{itemize}

\subsection{Implementation details of baseline model for 3D Image Segmentation}

The model is trained using $2\times$ NVIDIA A100 40GB GPUs. As input, we use a patch of $96 \times 96 \times 96$ with a batch size of $4$. We employ an AdamW optimizer with an initial learning rate of \num{1e-4} and weight decay of \num{1e-5} to minimize the loss function $\mathcal{L}$, which is defined as:
\begin{equation}
    \mathcal{L} = \mathcal{L}_{D} + \Lambda \cdot \mathcal{L}_{CE},
\end{equation}
\noindent
where $\mathcal{L}_{D}$, $\mathcal{L}_{CE}$ are Dice and Cross-Entropy loss, respectively. A grid search optimization $\in$ [0.5, 1.0] was performed which estimated an optimal value $\Lambda = 1$.

We use a cosine annealing learning rate scheduler \cite{loshchilov2017sgdr}. We implemented our network in Python v3.9 using PyTorch v2.1 and the MONAI library \cite{cardoso2022monai}. As an evaluation metric, we use the Dice Similarity Coefficient (DSC). We employ a one-way analysis of variance to evaluate the significance of variations among the segmentation performance metrics. We use $p < 0.05$ to distinguish a statistically significant difference.

\clearpage
\section{Use-case: explanatory analysis of a model for segmenting anatomical structures in CT scans} \label{app:explanations}

\begin{figure*}[ht]
    \centering
    \includegraphics[width=0.49\textwidth]{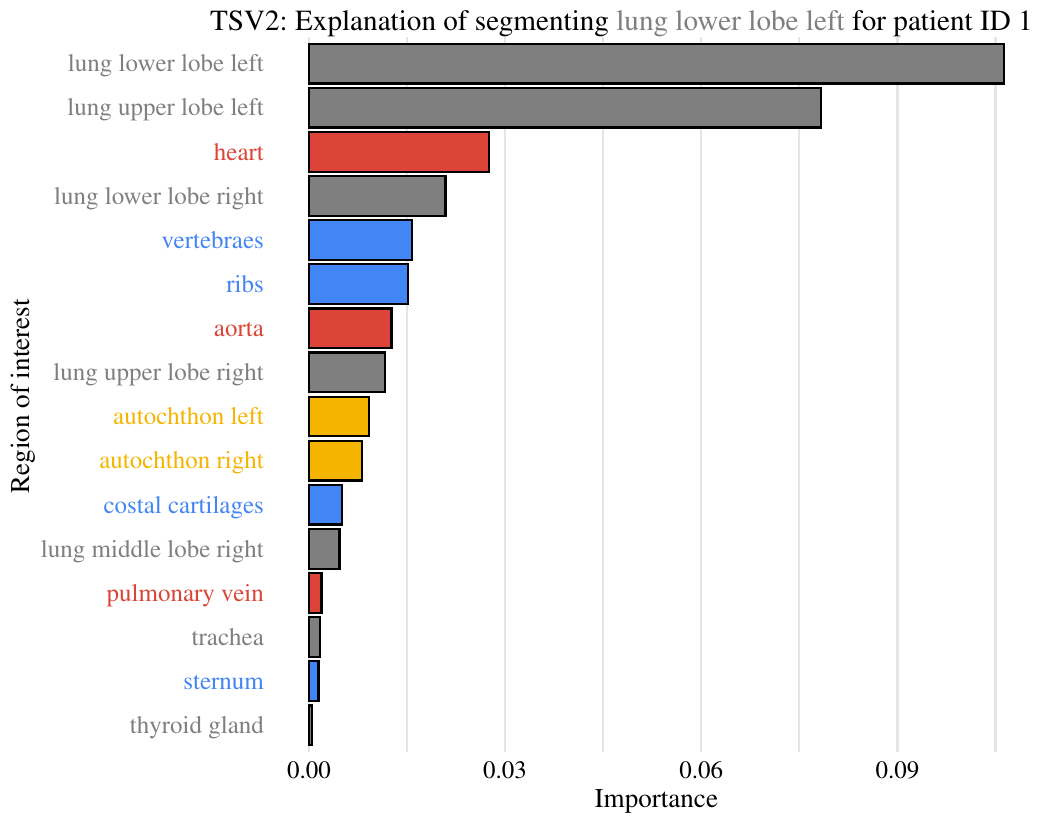}
    \includegraphics[width=0.49\textwidth]{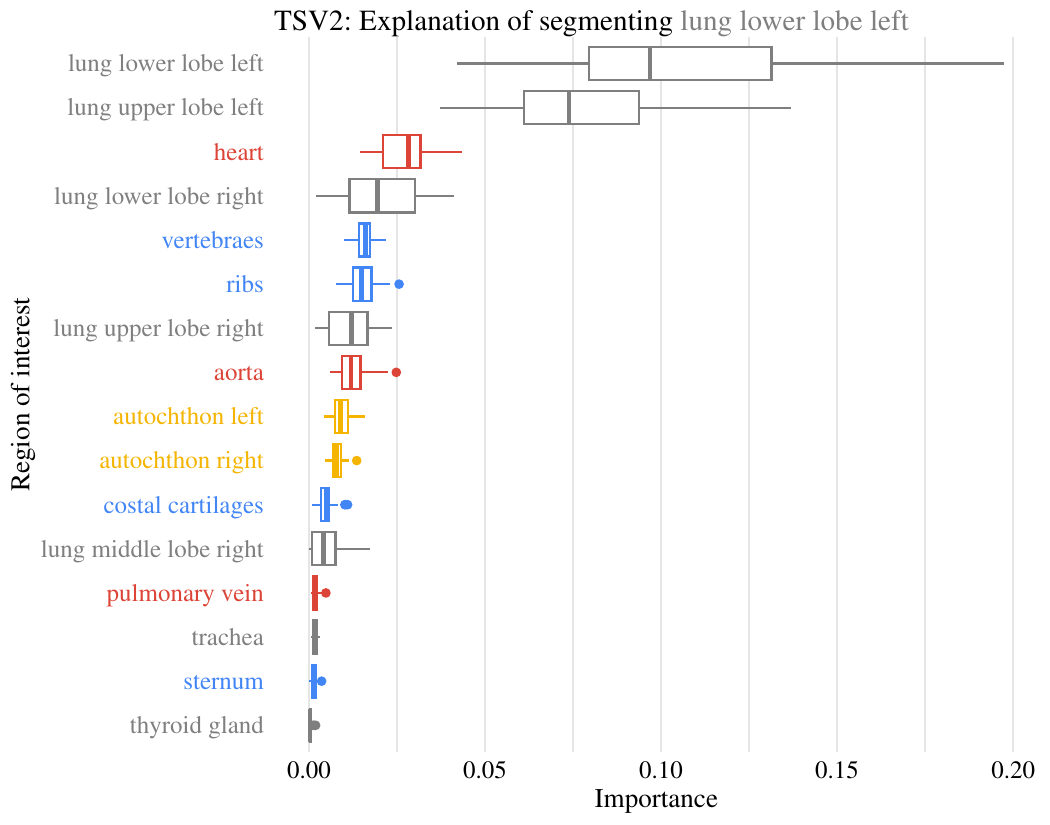}
    \includegraphics[width=0.49\textwidth]{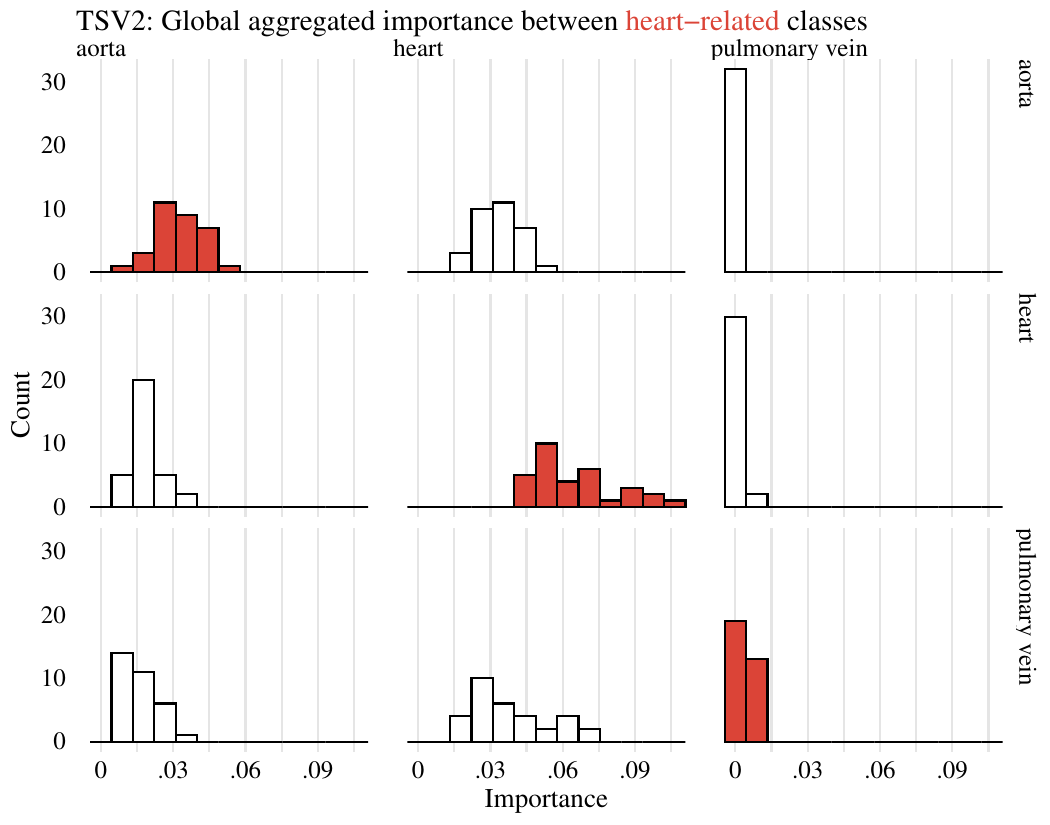}
    \includegraphics[width=0.49\textwidth]{figures/agg2exp_graph}
    \caption{\aggexp aggregates attributions into a global RoI importances denoting what a 3D segmentation model has learned. \textbf{Top-left:}~We aggregate local voxel attributions into local RoI importance for a single input (patient) and a single output class (\emph{lung lower lobe left}). \textbf{Top-right:}~We aggregate local RoI importances for a subset of inputs (patients) into global RoI importance.
    \textbf{Bottom-left:}~Global analysis of RoI importances for a subset of inputs between pairs of heart-related class labels. 
    \textbf{Bottom-right:}~\aggexp allows to discover a higher-level representation acquired by the complex segmentation model. Explanations are aggregated twice to obtain global importance between output class labels and other potential RoIs, e.g. lung pathologies.} 
    \label{fig:agg2exp}
\end{figure*}

\begin{figure*}[ht]
    \centering
    \includegraphics[width=0.49\textwidth]{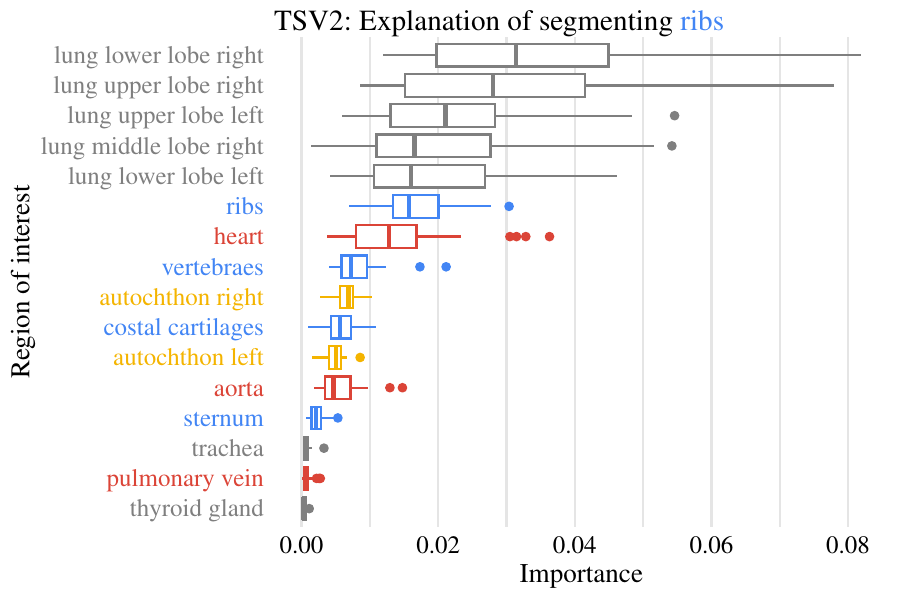}
    \includegraphics[width=0.49\textwidth]{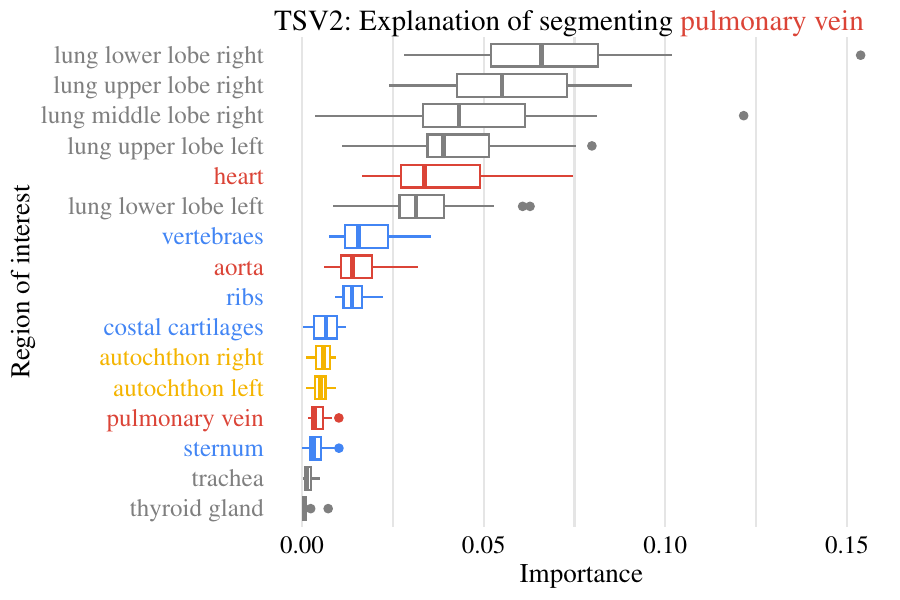}
    \caption{Distribution of local RoI importances for two class labels: \emph{ribs} and \emph{pulmonary vein}. We color objects by their semantic meaning, i.e., cardiovascular system in {\color{myred} \textbf{red}}, muscles in {\color{myyellow} \textbf{yellow}}, bones in {\color{myblue} \textbf{blue}}, other organs in {\color{darkgray} \textbf{grey}}.} 
    \label{fig:global_tsv2}
\end{figure*}

\begin{figure*}[ht]
    \centering
    \includegraphics[width=0.75\textwidth]{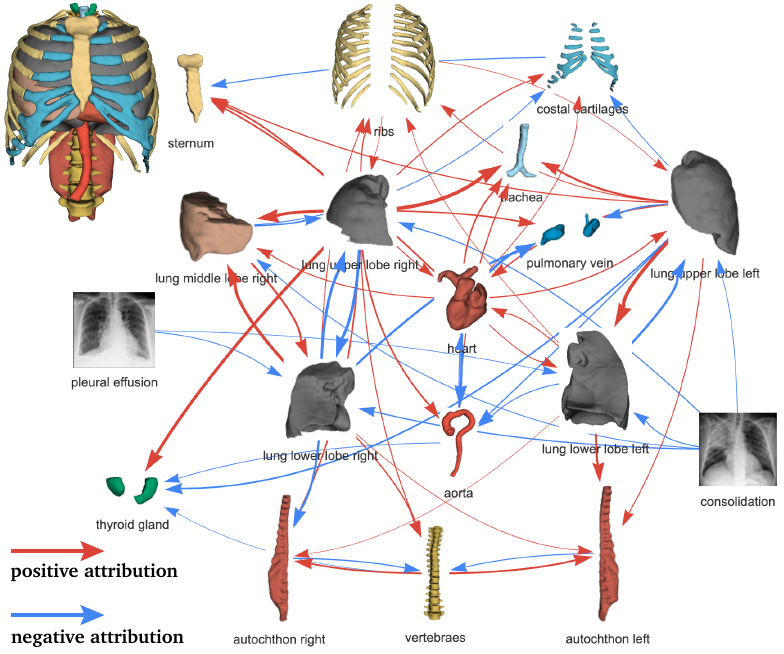}
    \caption{A global RoI importance explanation of the model's 3D segmentation. Aggregated attributions provide a measure of semantic importance between the segmented objects in B50. We visualize about one-third of the most important edges for clarity. The X-rays portraying the regions of lung pathologies are for illustration purposes only.} 
    \label{fig:graph_b50}
\end{figure*}

Here, we provide supplementary visualizations of explanations computed for both datasets. 
In \cref{fig:agg2exp}, we summarize the \aggexp methodology with four explanation visualizations.
\cref{fig:global_tsv2} shows the distribution of local RoI importances for TSV2; analogous to \cref{fig:glocal_b50} for B50. 
\cref{fig:graph_b50} shows the global RoI importance on a graph for B50 analogous to \cref{fig:graph_tsv2} for TSV2. 
Note that there are no annotations of lung pathologies available in the TSV2 dataset. 
We acknowledge that explanations computed on the external B50 dataset have more outliers, which seems reasonable as the model was trained on TSV2.
Moreover, especially comparing \cref{fig:graph_b50} to \cref{fig:agg2exp} (\textbf{bottom-right}), we observe differences in how the model utilizes the contextual physical and semantical information between the two validation datasets.

\clearpage
\subsection{Analysis of outliers with \aggexp}\label{app:outlier}

Tests' p-values and correlation coefficients are available in~\cref{tab:outlier_spearman_test}. When training Isolation Forest on all labels, we used 100 estimators. Each IF model was trained on the TSV2 train set that includes 490 CT scans.
\begin{table}[ht!]
    \centering
    \caption{P-values and correlation coefficients from Spearman correlation test between DSC and anomaly scores of the IF models.}
    \label{tab:outlier_spearman_test}
    \begin{tabular}{lrr}
        \toprule
        \textbf{Label} &   \textbf{p-value} & \textbf{Spearman Correlation} \\
        \midrule
        lung lower lobe right  &  0.000194 &     0.613 \\
        lung upper lobe right  &  0.000354 &     0.592 \\
        background             &  0.000432 &     0.585 \\
        costal cartilages      &   0.00431 &     0.491 \\
        lung middle lobe right &    0.0083 &     0.459 \\
        heart                  &    0.0181 &     0.415 \\
        sternum                &    0.0324 &     0.379 \\
        autochthon left        &    0.0399 &     0.365 \\
        ribs                   &     0.129 &     0.274 \\
        lung upper lobe left   &      0.13 &     0.273 \\
        autochthon right       &     0.139 &     0.267 \\
        trachea                &     0.145 &     0.264 \\
        lung lower lobe left   &     0.306 &     0.187 \\
        aorta                  &     0.446 &     $-$0.14 \\
        thyroid gland          &     0.463 &    $-$0.135 \\
        pulmonary vein         &      0.55 &      0.11 \\
        vertebraes             &     0.842 &    0.0367 \\
        \bottomrule
    \end{tabular}
\end{table}
\section{Comparison of the segmentation performance between the baseline models}

\bgroup
\def\arraystretch{1.5}%
\begin{table}[ht!]
    \caption{TotalSegmentator test set performance comparison of mean DSC for the following classes. For better readability, we present the left and right lungs as mean of their subparts and autochthon as one class: Aorta (A), Left Lung (LL), Right Lung (RL), Trachea (T), Heart (H), Pulmonary Vein (PV), Thyroid Gland (TG), Ribs (R), Vertebrae (V), Autochthon (AU), Sternum (S), Costal Cartilages (CC). *indicates statistical significance between Swin UNETRv2 and other state-of-the-art methods mDSC ($p < 0.05$). **indicates training with self-supervised pre-trained weights.}
    \begin{center}
    \resizebox{\textwidth}{!}{%
        \begin{tabular}{@{}l|cccccccccccc|cc@{}}
        \toprule
        \textbf{Method} & \textbf{A} & \textbf{LL} & \textbf{RL} & \textbf{T} & \textbf{H} & \textbf{PV} & \textbf{TG} & \textbf{R} & \textbf{V} & \textbf{AU} & \textbf{S} & \textbf{CC} & \textbf{mDSC $\uparrow$} \\
        \midrule
        Swin UNETRv1** \cite{tang2022self} & 78.02 & 85.50 & 83.93 & 85.36& 87.93& 69.26& 65.23 & 82.07& 84.68& 83.20 & 63.57& 75.42& 82.16 (*) \\
        3D U-Net \cite{cciccek20163d} & 86.35 & 90.80 & 88.26 & 85.15 & 90.57 & 71.91 & 71.81 & 89.19 & 86.56 & 88.44 & 81.73 & 78.65 & 87.11 (*)\\
        UNETR \cite{hatamizadeh2022unetr} & 87.23 & 90.27 & 87.94 & 87.20 & 90.01 & 74.65 & 72.85 & 91.90 & 89.65 & 85.45 & 81.70 & 81.48 & 87.41 (*) \\
        Swin UNETRv1 \cite{tang2022self} & 89.13 & 91.75 & 88.62 & 89.27& 90.08 & 78.51 & 78.88 & 92.03 & 92.25 & 90.32 & 80.74 & 85.13 & 89.38 (*) \\
        Swin UNETRv2 \cite{he2023swinunetr} & 89.15 & 92.80 & 90.34 & 89.17 & 91.18 & 77.81 & 80.41 & 92.28 & 92.65 & 91.35 & 83.16 & 85.80 & \textbf{90.24} \\
        \bottomrule
        \end{tabular}
        }
    \end{center}
    \label{tab:segmentationresults}
\end{table}
\egroup

\end{document}